\documentclass[]{oppo_arxiv}

\usepackage[toc,page,header]{appendix}
\usepackage{fancyvrb}
\usepackage{wrapfig}
\usepackage[authoryear]{natbib}
\usepackage[percent]{overpic}
\usepackage{algorithm}
\usepackage{algorithmic}
\usepackage{tikz}
\usetikzlibrary{positioning, calc}

\usepackage{times}
\usepackage{latexsym}
\usepackage[T1]{fontenc}
\usepackage[utf8]{inputenc}
\usepackage{microtype}
\usepackage{inconsolata}
\usepackage{graphicx}
\usepackage{hyperref}       
\usepackage{url}            
\usepackage{booktabs}       
\usepackage{amsfonts}       
\usepackage{nicefrac}       
\usepackage{stackengine}
\usepackage{microtype}      
\usepackage{colortbl}
\usepackage{xcolor}
\usepackage{amsmath}
\usepackage{amssymb}
\usepackage{amsthm}
\usepackage{mathrsfs}
\usepackage{pifont}
\usepackage{MnSymbol}
\usepackage{balance}
\usepackage{enumitem}
\usepackage{xcolor}
\usepackage{natbib}
\usepackage{multicol}
\usepackage{xspace}
\usepackage{fancyvrb}
\usepackage{geometry}
\usepackage{tocloft}
\usepackage{tasks}
\usepackage{hyperref}
\usepackage{fontawesome5}
\usepackage{makecell}
\AtBeginDocument{%
  \providecommand\BibTeX{{%
    \normalfont B\kern-0.5em{\scshape i\kern-0.25em b}\kern-0.8em\TeX}}}

\makeatletter
\DeclareRobustCommand\onedot{\futurelet\@let@token\@onedot}
\def\@onedot{\ifx\@let@token.\else.\null\fi}

\usepackage{setspace}
\usepackage{mathtools}

\usepackage{multirow,booktabs}
\usepackage{subcaption}

\title{Search More, Think Less: Rethinking Long-Horizon Agentic Search for Efficiency and Generalization}

\affiliation{OPPO AI Agent Team}

\abstract{
Recent deep research agents primarily improve performance by scaling reasoning depth, but this leads to high inference cost and latency in search-intensive scenarios. Moreover, generalization across heterogeneous research settings remains challenging. In this work, we propose \emph{Search More, Think Less} (SMTL), a framework for long-horizon agentic search that targets both efficiency and generalization. SMTL replaces sequential reasoning with parallel evidence acquisition, enabling efficient context management under constrained context budgets. To support generalization across task types, we further introduce a unified data synthesis pipeline that constructs search tasks spanning both deterministic question answering and open-ended research scenarios with task appropriate evaluation metrics. We train an end-to-end agent using supervised fine-tuning and reinforcement learning, achieving strong and often state of the art performance across benchmarks including BrowseComp (48.6\%), GAIA (75.7\%), Xbench (82.0\%), and DeepResearch Bench (45.9\%). Compared to Mirothinker-v1.0, SMTL with maximum 100 interaction steps reduces the average number of reasoning steps on BrowseComp by 70.7\%, while improving accuracy.

}

\date{\today}
\checkdata[Open Source]{%
  \href{\detokenize{https://github.com/OPPO-PersonalAI/Agent_Foundation_Models}}{%
    \faGithub\ Code%
  }\quad
  \href{\detokenize{https://huggingface.co/PersonalAILab/SMTL-30B}}{%
    \raisebox{-.15em}{\includegraphics[height=1em]{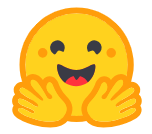}} Models%
  }\quad
  \href{\detokenize{https://huggingface.co/datasets/PersonalAILab/SMTL-Dataset}}{%
    \faDatabase\ Datasets%
  }%
}

\correspondence{Wangchunshu Zhou at \email{zhouwangchunshu@oppo.com}}

\begin{document}
\maketitle

\begin{figure*}[h]
  \centering
  \begin{subfigure}[t]{0.48\linewidth}  
    \centering
    \includegraphics[width=\linewidth]{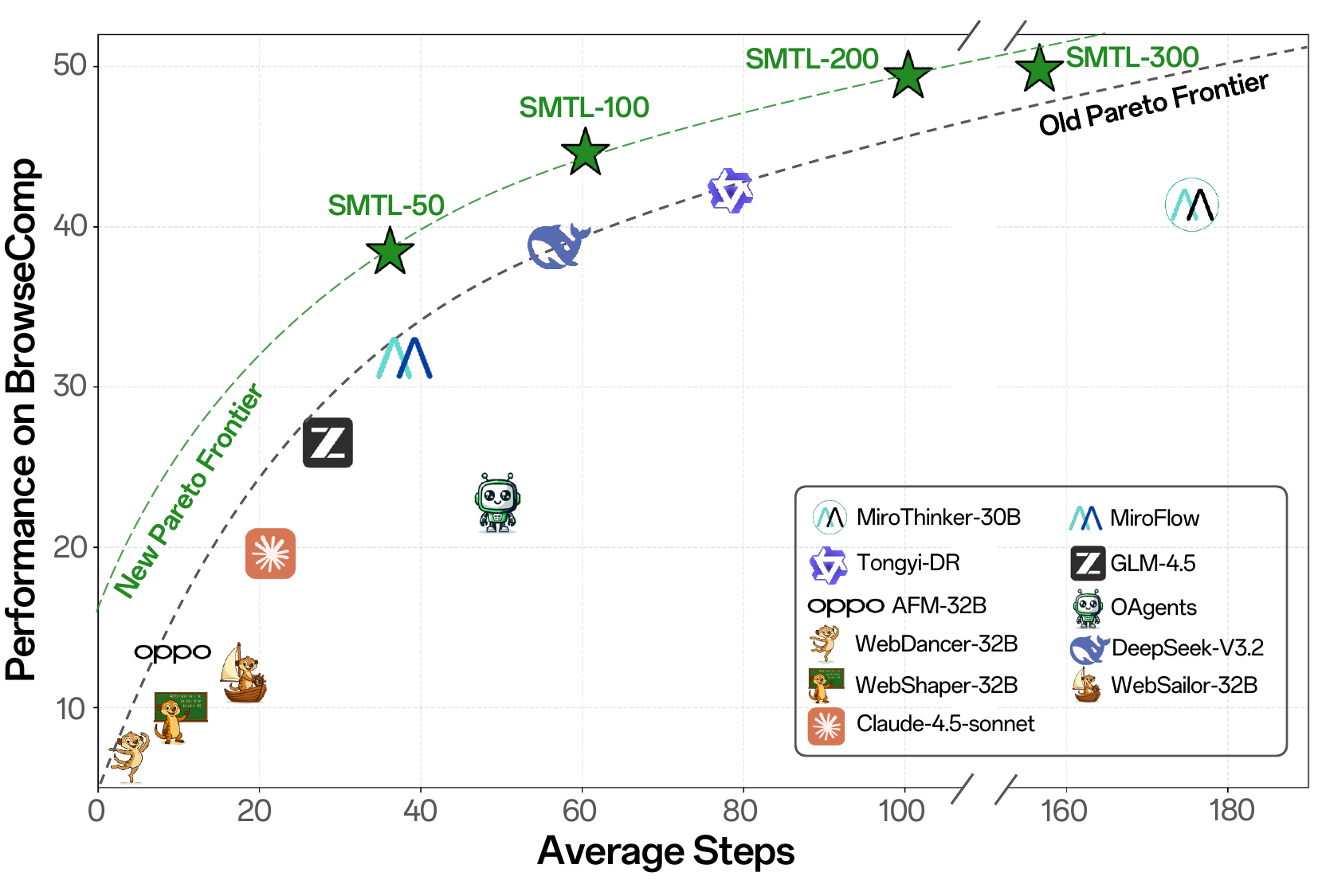}
    \vspace{-15pt}
    \caption{Efficiency on BrowseComp: score vs. interaction steps.}
    \label{fig:efficiency}
  \end{subfigure}
  \hfill
  \begin{subfigure}[t]{0.48\linewidth}  
    \centering
    \includegraphics[width=\linewidth]{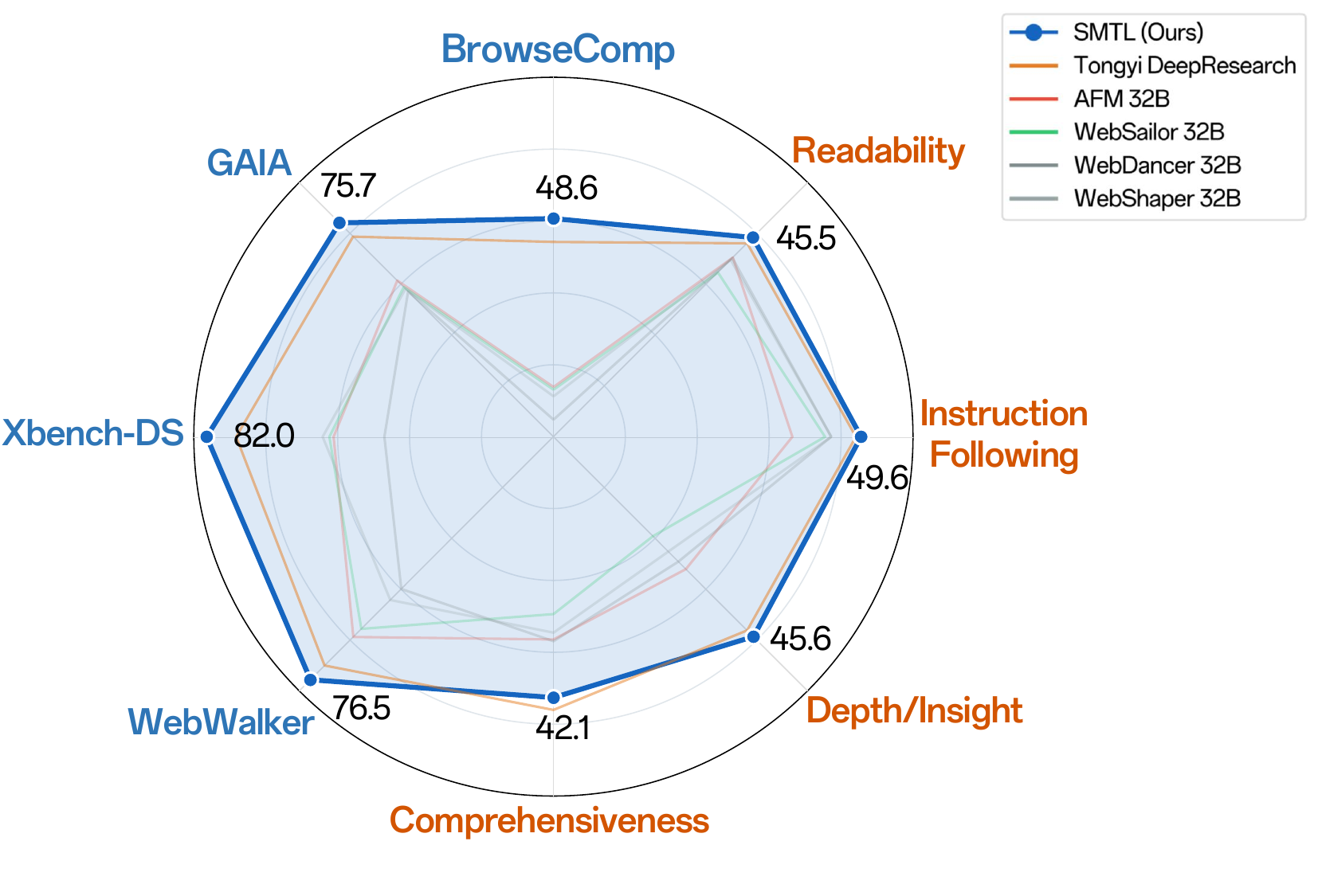}
    \vspace{-15pt}
    \caption{Generalization across multiple benchmarks.}
    \label{fig:generalization}
 \end{subfigure}
   \caption{\textbf{Overview of SMTL Performance.} (a) Efficiency on BrowseComp. All methods are evaluated with their default settings. (b) Generalization across benchmarks. Comprehensiveness, depth, instruction following, and readability are measured following DeepResearch Bench. We report SMTL-300 for Deep Search benchmarks and SMTL-100 for Deep Research benchmarks.}
  \label{fig:overview}
\end{figure*}

\clearpage
\setcounter{tocdepth}{3}

\renewcommand{\contentsname}{\LARGE\bfseries Contents}

\renewcommand{\cftsecfont}{\large\bfseries}
\renewcommand{\cftsecpagefont}{\large\bfseries}

\renewcommand{\cftsubsecfont}{\normalfont\itshape\normalsize}
\renewcommand{\cftsubsecpagefont}{\normalfont\itshape\normalsize}

\renewcommand{\cftsubsubsecfont}{\normalfont\itshape\normalsize}
\renewcommand{\cftsubsubsecpagefont}{\normalfont\itshape\normalsize}

\setlength{\cftbeforesecskip}{5pt}
\setlength{\cftbeforesubsecskip}{3pt}
\setlength{\cftbeforesubsubsecskip}{2pt}


\section{Introduction}
Recent advances in deep research agents suggest that increasing reasoning depth and the number of tool calls can substantially improve task performance~\citep{jin2025search, li2025search, li2025webthinker, wu2025webdancer, sun2025simpledeepsearcher, zhang2025evolvesearch, zheng2025deepresearcher, zhou2023agents, zhou2024agents2, zhu2025oagentsempiricalstudybuilding, zhu2025scalingtesttimecomputellm, qiu2025alita, smolagents, tang2025agent}. However, longer reasoning trajectories also lead to higher inference latency. Balancing long-horizon search performance and computational efficiency remains an open problem.~\citep{bai2025mirothinker, li2025tongyi, chen2025minimax, tang2025deepminer}. To train agent models that can perform efficient long-horizon reasoning, a key question is:
How can we design tasks that require efficient long-horizon reasoning, and construct suitable training trajectories to teach such behavior?

In addition, generalization across diverse task objectives and evaluation criteria remains challenging. Existing agentic search tasks can be roughly divided into two types. The first includes deterministic question-answering tasks with clear ground-truth answers, such as BrowseComp~\citep{wei2025browsecomp}, GAIA~\citep{mialon2023gaia}, and WebWalker~\citep{wu2025webwalker}, where performance is mainly measured by accuracy. The second type focuses on open-ended research problems without a single correct answer, such as DeepResearch Bench~\citep{du2025deepresearchbench} and DeepResearchGym~\citep{coelho2025deepresearchgym}, where evaluation emphasizes information coverage, coherence, and synthesis quality. These two settings have very different optimization objectives. As a result, agents trained for one setting may struggle to generalize to the other, making it difficult to train a single agent that performs well across both.

In this work, we revisit long-horizon agentic search from the perspectives of efficiency and generalization. We argue that the main scalability bottleneck of existing deep research agents lies in their reliance on linear, sequential reasoning in search tasks. To address this, we propose an agentic framework that replaces sequential reasoning with parallel task decomposition and concurrent tool execution, together with structured context management for efficient long-horizon inference under constrained context budgets. At the data level, we introduce an automated pipeline that synthesizes representative multi-type search tasks, reducing redundant samples and improving generalization.
Trained end-to-end with supervised fine-tuning and reinforcement learning, our approach achieves state-of-the-art performance on BrowseComp (44.3), while substantially improving efficiency by reducing the number of reasoning steps by 78\% and inference latency by up to 2.6$\times$.

Our contributions are summarized as follows:
\begin{itemize}[leftmargin=12pt, itemsep=-3pt, topsep=-3pt]
    \item \textbf{Parallel agentic workflow.} We propose a unified agentic framework that replaces sequential reasoning with parallel evidence acquisition, using plan-driven context management to achieve efficient long-horizon search under constrained context budgets.

    \item \textbf{Generalized data construction.} We introduce an automated data pipeline that constructs representative multi-type search tasks across both deterministic and open-ended settings, supporting generalization in long-horizon agentic search.

    \item \textbf{State-of-the-art performance.} 
    Our method achieves state-of-the-art results on multiple deep search and deep research benchmarks, while substantially improving efficiency in terms of reasoning steps and inference latency.


\end{itemize}
\section{Related Work}
\paragraph{Agent Frameworks and Systems.}
Agentic workflows augment large language models (LLMs) with planning, multi-step tool use, and iterative environment interaction, and have become the dominant paradigm for search-intensive tasks. A mainstream line of work relies on external orchestration, where a controller decomposes queries into subgoals, schedules web search, browsing, and code-execution tools, and aggregates intermediate findings through predefined procedures. This paradigm underlies recent commercial deep research systems that combine strong proprietary backbones with multi-step web exploration, plan refinement, and long-context memory to maintain intermediate evidence \citep{openai_deep_research_2025,google_gemini_deep_research_2024,anthropic_claude_research_2025,perplexity_deep_research_2025}. In parallel, structured agentic frameworks such as WebWeaver \citep{li2025webweaver} and OAgents \citep{zhu2025oagentsempiricalstudybuilding} adopt planner--researcher or planner--executor workflows to improve robustness in open-ended research and verification tasks. The open-source ecosystem further provides reusable orchestration templates, including plan-and-execute pipelines and hierarchical agent systems \citep{langchain_open_deep_research_2025,skyworkai_deepresearchagent_2025,togetherai_open_deep_research_2025}, with MiroFlow consolidating these patterns into a benchmark-oriented research-agent framework \citep{miromindai_miroflow_2025}. Multi-agent workflows additionally introduce division of labor through specialized roles and coordination cycles \citep{li2023camel,hong2024metagpt,qian2023chatdev,fourney2024magentic,smolagents,chen2023agentverse,hu2025owl}. 
Despite architectural diversity, most existing agentic workflows implicitly scale performance by deepening sequential reasoning and expanding interaction horizons. As a consequence, this paradigm often leads to limited information efficiency, as substantial computation is devoted to prolonged model-side reasoning rather than effective external evidence acquisition.

\paragraph{Synthetic Data Pipelines.}
High-quality training data is critical for search agents, yet collecting multi-step, tool-interactive trajectories at scale remains costly. Recent synthetic data pipelines can be broadly categorized into two approaches. The first is graph-based generation, exemplified by WebSailor~\citep{li2025websailor}, which constructs knowledge graphs from seed entities using web tools and samples complex question--answer pairs or trajectories from subgraphs. The second approach follows an easy-to-hard expansion paradigm, where simple seed questions are progressively expanded into longer-horizon problems with predominantly tree-structured logic, as seen in WebShaper, ASearcher, and WebExplorer \citep{tao2025webshaper,gao2025beyond,liu2025webexplorer}. Beyond web-centric pipelines, TaskCraft expands atomic tasks along both depth and width dimensions and applies incremental validation mechanisms, thereby improving data quality and controllability \citep{shi2025taskcraft}. Overall, existing synthetic pipelines highlight the effectiveness of tool-in-the-loop generation, while primarily emphasizing task difficulty or context length rather than explicitly shaping information-efficient search and verification behaviors. Besides, these pipelines are primarily designed around deterministic question answering or tightly constrained task structures, and provide limited support for synthesizing open-ended research tasks that require flexible information aggregation and cross-source validation.

\section{Parallel Agentic Workflow}
\label{sec_workflow}
\begin{figure}[!ht]
    \centering
    \includegraphics[width=\linewidth]{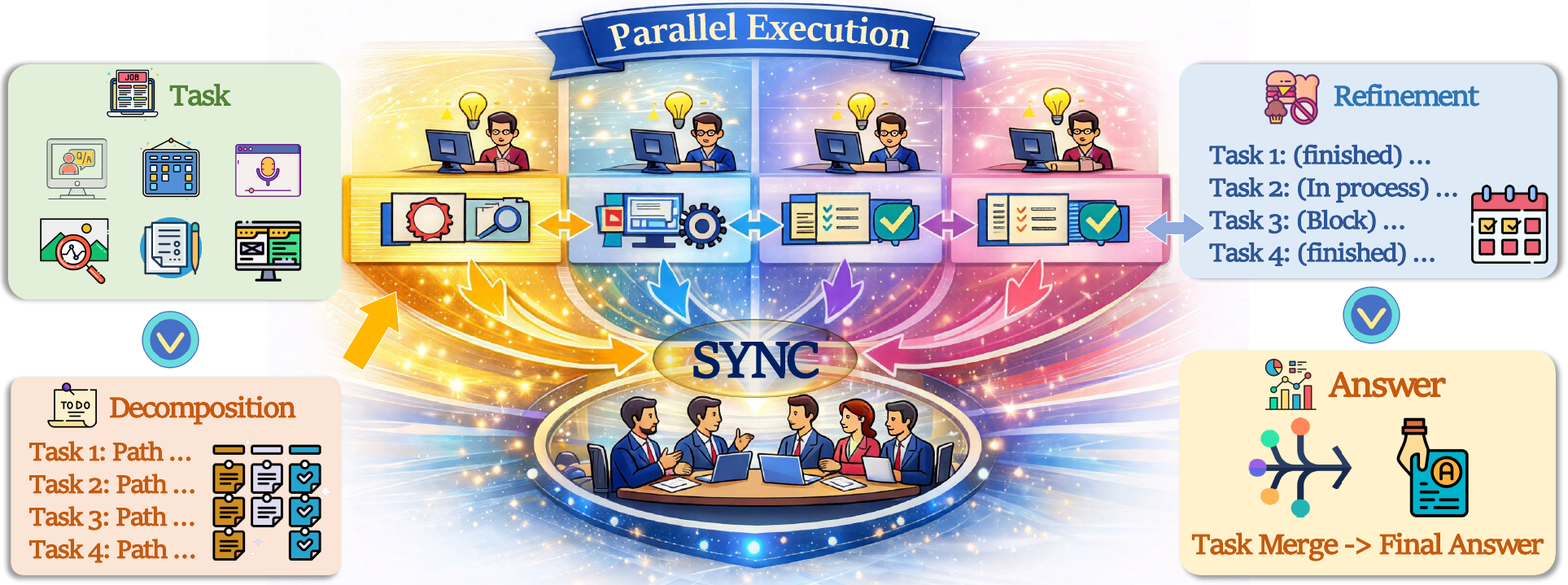}
    \vspace{-15pt}
    \caption{Overview of our parallel agentic workflow design.}
    \label{fig:workflow}
\end{figure}

Recent tool-augmented agents enable language models to interact with external tools for retrieval and reasoning~\citep{yao2023react,xue2025simpletir,li2025webthinker}. 
Building on this line of work, we design an efficient parallel agentic workflow inspired by Flash-Searcher~\citep{qin2025flash}, which explicitly supports concurrent execution and structured coordination across subtasks. As illustrated in \Cref{fig:workflow}, the agent initializes a task plan, executes multiple subtasks in parallel, and periodically synchronizes intermediate results through plan refinement before producing the final answer.

\paragraph{Initial Plan Construction.}
Given a composite search task, the agent first constructs an initial task plan \( G_{\text{plan}}^{0} \) by decomposing the problem into a set of interrelated yet partially independent subtasks. Each subtask corresponds to a concrete information-seeking or verification objective, such as retrieving facts, validating relations, or gathering evidence. The plan is generated prior to any tool execution and is designed to expose parallelizable execution paths early, enabling concurrent evidence acquisition and higher information density. Rather than relying on a single sequential reasoning chain, \( G_{\text{plan}}^{0} \) provides a structured starting point that supports parallel execution and subsequent refinement.

\paragraph{Parallel Execution and Tool Coordination.}
At each timestep, the system selects ready-to-execute subtasks from the pending set \( \mathcal{P}_t \) based on their readiness, while completed subtasks are tracked separately. These pending subtasks are processed concurrently, leveraging available tools or agent actions to gather information and execute reasoning tasks. By executing multiple pending subtasks in parallel, the system accelerates task completion and reduces sequential bottlenecks. The system aggregates observations from each parallel execution into a unified reasoning state:
\begin{equation}
s_{t+1} = \mathcal{F}(s_t, \{a^{(k)}_t\}_{k=1}^{m}, \{o^{(k)}_t\}_{k=1}^{m}),
\end{equation}
where \( a^{(k)}_t \) and \( o^{(k)}_t \) represent the actions and observations of the \( k \)-th parallel execution, and \( s_t \) denotes the aggregated reasoning state at timestep \( t \).

In practice, parallel execution is realized via a limited set of reusable external tools (Appendix~\ref{app:tools}), including web search and page crawling, which are repeatedly invoked across pending subtasks to facilitate concurrent information acquisition and verification.

\paragraph{Dynamic Plan Refinement.}
To ensure the plan adapts to the ongoing execution, the task plan is periodically updated. Completed subtasks are removed, unresolved dependencies are rechecked, and new subtasks may be introduced. The task plan is refined based on the current execution state:
\[
G_{\text{plan}}^{t+\Delta} = \mathcal{R}(G_{\text{plan}}^t, \mathcal{C}_t, \mathcal{P}_t, s_t),
\]
where \( \mathcal{C}_t \) represents the completed subtasks. This dynamic refinement ensures that the task adapts to progress and maintains efficiency.

\begin{algorithm}[!ht]
\caption{Parallel agentic workflow.}
\begin{algorithmic}[1]
\REQUIRE Composite task \( T \)
\STATE Construct initial task plan \( G_{\text{plan}}^{0} \gets \mathcal{D}(T) \)
\STATE Initialize reasoning state \( s_0 \), pending subtasks \( \mathcal{P}_0 \gets G_{\text{plan}}^{0} \), completed subtasks \( \mathcal{C}_0 \gets \emptyset \)
\WHILE{\( \mathcal{P}_t \neq \emptyset \)}
    \STATE Select executable subtasks \( \mathcal{E}_t \subseteq \mathcal{P}_t \)
    \STATE Execute subtasks in \( \mathcal{E}_t \) in parallel using available tools
    \STATE Update reasoning state \( s_{t+1} = \mathcal{F}(s_t, \{a^{(k)}_t\}, \{o^{(k)}_t\}) \)
    \STATE Mark completed subtasks and update \( \mathcal{C}_{t+1} \)
    \STATE Remove completed subtasks from \( \mathcal{P}_{t+1} \)
    \IF{\( t \bmod \Delta = 0 \)}
        \STATE Refine task plan using current execution state
        \STATE \( G_{\text{plan}}^{t+1} \gets \mathcal{R}(G_{\text{plan}}^{t}, \mathcal{C}_{t+1}, \mathcal{P}_{t+1}, s_{t+1}) \)
    \ENDIF
\ENDWHILE
\STATE \textbf{Return} final reasoning state \( s_T \)
\end{algorithmic}
\label{alg:pipeline}
\end{algorithm}

\paragraph{Algorithmic Outline.}
The workflow design is summarized in Algorithm~\ref{alg:pipeline}. This method continuously refines the task plan while executing multiple subtasks in parallel, ensuring efficient task completion through dynamic updates and parallel execution.

\section{Data Construction}
\label{sec_data}

Existing agent data construction pipelines limit both generalization and efficiency. Many rely on static knowledge sources and focus on deterministic, entity-centric Deep Search tasks, providing weak coverage of open-ended Deep Research scenarios. Moreover, task difficulty is often scaled by increasing reasoning hops rather than improving information density, leading to redundant evidence and inefficient interaction traces.

\begin{figure}[!ht] 
    \centering 
    \includegraphics[width=\linewidth]{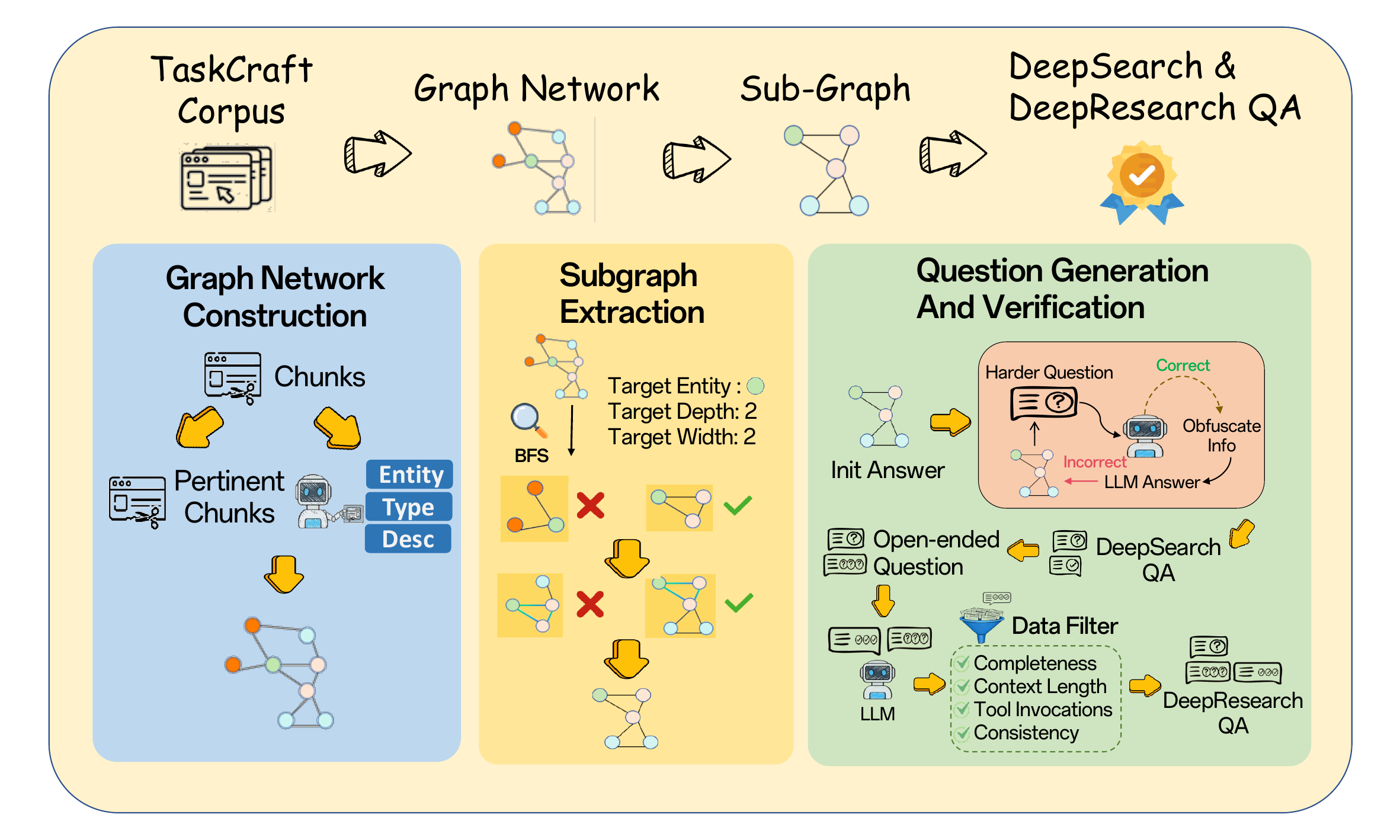} 
    \vspace{-18pt}
    \caption{Overview of the data construction pipeline.  
    }
    \label{fig:data synthesis} 
\end{figure}

To address these issues, we propose a \textbf{high-diversity, high-density} data synthesis pipeline. As illustrated in Fig.~\ref{fig:data synthesis}, our framework integrates raw corpus collection, graph construction, subgraph extraction, and QA generation with verification, enabling dense and correlated evidence acquisition for both deterministic and open-ended research tasks.

\subsection{Deep Search Data}

\paragraph{Raw Corpus Collection.}
To support cross-domain generalization and multi-hop reasoning, we construct the raw corpus by leveraging trajectories from TaskCraft corpus~\citep{shi2025taskcraft}. These trajectories contain rich collections of real-world URLs spanning diverse domains, including art, sports, history, government, economics, politics, music, geography, movies, computer science, physics, and chemistry. 

Crucially, URLs within each trajectory are not independent: they are connected through explicit information-seeking paths, where later queries and sources build upon evidence gathered from earlier ones. This structure naturally induces multi-hop relationships across documents, making the collected corpus well suited for graph-based task construction. We extract and normalize the content from these URLs using document parsing tool Jina \citep{JinaAI_2025_JinaReader}, yielding a large-scale, high-diversity raw corpus that preserves both domain coverage and inter-document relational structure. 


\paragraph{Graph Network Construction.}
Based on the initial corpus, we develop an efficient pipeline for generating sophisticated graph networks. First, we leverage LightRAG \citep{guo2024lightrag} to instantiate a knowledge-driven graph network. This process involves partitioning the curated text crops into multiple chunks, from which an LLM extracts entities and their respective attributes. By integrating an embedding-plus-reranker retrieval mechanism, we recall pertinent chunks, enabling the LLM to synthesize detailed node descriptions and delineate complex inter-node relationships, ultimately culminating in a highly intricate graph network.

\paragraph{Subgraph Extraction.}
Given the constructed knowledge graph, we extract task-specific subgraphs using a controlled random-walk strategy. For each task, we sample a target entity as the ground-truth answer and perform a breadth-first search (BFS) up to $N$ hops to collect its surrounding neighborhood. The resulting subgraph defines the supporting evidence structure required to infer the answer, where multi-hop nodes serve as question conditions with varying degrees of indirection. By adjusting the hop depth and branching factor, we flexibly control task difficulty while preserving semantic coherence and factual correctness. This construction yields compact, information-dense task skeletons that require integrating multiple related evidence sources rather than relying on single-hop retrieval.

To ensure high-quality and non-degenerate task structures, we adopt the following design principles:

\begin{itemize}[leftmargin=8pt, itemsep=-3pt, topsep=-3pt]
\item \textbf{Rich topological structures.} We prioritize subgraphs in which two $(N{+}1)$-hop nodes are interrelated while sharing a common $N$-hop parent, inducing cyclic dependencies that require cross-validation of multiple relationships.
\item \textbf{Controlled walk width and depth.} We explicitly bound the depth and branching factor to keep task difficulty scalable and avoid trivial shortcuts or excessively long reasoning chains.
\end{itemize}

\paragraph{Hierarchical Question Construction and Verification.}
Given a task-specific subgraph with a fixed target answer, we construct questions through a hierarchical synthesis process. Starting from the outermost $N$-hop frontier, we iteratively aggregate information from $(i{+}1)$-hop nodes to form sub-questions about $i$-hop entities. Each aggregation step yields a valid intermediate question, and progressively merging all layers produces a final question about the target entity that requires the maximum hop depth and reasoning difficulty.

When multiple $(i{+}1)$-hop nodes exhibit semantic relationships, we explicitly encode these interdependencies as verifiable conditions, requiring agents to cross-validate parallel evidence paths rather than rely on linear reasoning. To prevent information leakage, we apply an LLM-based verification step after each synthesis iteration; if the answer can be inferred prematurely, the question is restructured or relevant information is obfuscated. This process repeats until the desired difficulty is achieved or a maximum of five iterations is reached.

The complete designs of the prompts for entity information extraction, entity evaluation, entity question generation and obfuscation, and verification are described in~\Cref{app:sec:search_data_construct}.

\subsection{Deep Research Data}

\paragraph{Research Question Construction.}
Deep research tasks are synthesized entirely within our unified data construction pipeline, without relying on externally curated queries. Given a task-specific subgraph with a fixed target entity and its multi-hop supporting structure, we formulate open-ended research questions that require integrating evidence across the entire subgraph. These questions are designed to elicit report-style answers involving explanation, comparison, and synthesis across multiple sources, rather than single factual outputs. This construction explicitly encourages long-horizon planning, information aggregation, and cross-source reasoning.

\paragraph{Trajectory Generation and Quality Filtering.}
For each open-ended research question, we generate multiple candidate trajectories using our parallel agentic workflow and apply a two-stage filtering process to ensure high-quality supervision. First, trajectories are subjected to rule-based hard rejection to remove structurally invalid or overly shallow solutions, enforcing basic requirements on completeness, context validity, reasoning depth, tool usage, and format consistency. Second, trajectories that pass these checks are then evaluated by an LLM-as-a-Judge, which assesses higher-level semantic quality, including comprehensiveness, insight/depth, instruction following, and readability. Only trajectories that satisfy both structural constraints and semantic criteria are retained as reference outputs, yielding scalable and reliable training data for open-ended deep research tasks.

The complete designs of the prompts for open-ended question construction and quality filtering are illustrated in ~\Cref{app:sec:research_data_construct}.
\section{Training Recipe}

\subsection{Post-training Supervised Fine-tuning.}
We perform supervised fine-tuning (SFT) to initialize the agent with stable and efficient search behaviors before reinforcement learning.

\paragraph{Task Composition.}
The SFT dataset consists of two task categories: \emph{Deep Search} and \emph{Deep Research}, which differ in supervision form while sharing the same underlying subgraph-based construction.

\begin{itemize}[leftmargin=8pt, itemsep=-3pt, topsep=-3pt]
    \item \textbf{Deep Search.} Tasks are instantiated from task-specific subgraphs with hop depth ranging from 2 to 5. For each subgraph, all hierarchical question variants constructed during iterative aggregation are retained, yielding multiple questions that share the same target entity as the ground-truth answer. To prevent over-representation of frequent answers, we apply an answer frequency threshold and discard tasks whose target entities appear excessively often.

    \item \textbf{Deep Research.} For each subgraph, we construct an open-ended research question centered on the target entity and its multi-hop supporting structure. Questions are formulated to encourage broad exploration and synthesis over the entire subgraph, rather than single-answer retrieval, ensuring sufficient topical richness and variability.
\end{itemize}

\paragraph{Trajectory Construction and Curation.}
Training trajectories are generated using the agentic workflow described in Section~\ref{sec_workflow}. For Deep Search tasks, supervision is obtained by distilling trajectories generated by DeepSeek-V3.2 \citep{liu2025deepseekv32}, while Deep Research trajectories are distilled from GPT-5 \citep{openai2025gpt5}, reflecting its stronger long-form synthesis capabilities.

To ensure high-quality supervision, we apply the following curation criteria:
\begin{itemize}[leftmargin=8pt, itemsep=-3pt, topsep=-3pt]
    \item \textbf{Context length constraint.} The total trajectory length is capped at 64K tokens to reduce redundant interactions and noisy supervision.
    
    \item \textbf{Interaction efficiency constraint.} The average number of tool calls per step must be no less than 3, encouraging active information acquisition.
    
    \item \textbf{Trajectory efficiency optimization.} For tasks with multiple successful trajectories, we retain only those that are correct and shortest in interaction length.
\end{itemize}

\subsection{Reinforcement Learning.}

We adopt a slightly modified version of the REINFORCE Leave-One-Out (RLOO) algorithm \citep{ahmadian2024basicsrevisitingreinforcestyle} as our reinforcement learning method. Compared to GRPO, RLOO provides an unbiased advantage estimator. Our modifications are as follows. 
First, following the implementation in DAPO \citep{yu2025dapo}, we employ a token-level loss function. 
Second, to mitigate the training--inference mismatch arising from discrepancies between the inference engine and the training framework in log-probability computation, we apply sequence-level importance sampling for rollout correction \citep{liu-li-2025-rl-collapse}. 
Third, to ensure trajectory quality, we filter out certain negative trajectories so that they do not participate in advantage estimation or gradient updates. These negative trajectories include (i) failures caused by environmental issues such as connection timeouts or server errors, and (ii) responses that are excessively long or reach the maximum number of turns. This filtering strategy prevents the model from learning spurious behaviors induced by environmental instability and effectively stabilizes training.

During the RL stage, we optimize trajectories using an outcome-based reward. An LLM-as-a-judge evaluates whether the final answer is correct, assigning a reward of 1 for correct answers and 0 otherwise. Notably, if a tool call violates the required format, generation is immediately terminated and a reward of 0 is assigned, thereby explicitly encouraging correct tool usage.



\section{Experiments}
\subsection{Experimental Setup}

\paragraph{Baselines.}
For comparison, we conducted a comprehensive evaluation of our SMTL model against three categories of systems: (1) foundation models with tools, including closed-source models such as Claude-4.5-Sonnet~\cite{anthropic2025claude}, OpenAI-GPT-5~\cite{openai2025gpt5}, Gemini-2.5-Pro~\cite{comanici2025gemini25}, and open-source models GLM-4.5~\cite{zeng2025glm}, Minimax-M2~\cite{minimax2025minimaxm2}, DeepSeek-V3.2~\cite{liu2025deepseekv32}, Kimi-K2-0905~\cite{team2025kimi}; (2) deep research system, including OpenAI DeepResearch~\cite{openai_deep_research_2025}, Gemini DeepResearch~\cite{gemini_deep_research_2025}, Perplexity Deep Research~\cite{Perplexity2025}, Kimi-Researcher~\cite{moonshot2025kimiresearcher}, OAgents (Claude-3-7)~\cite{zhu2025oagentsempiricalstudybuilding}, MiroFlow (GPT-5)~\cite{miromindai_miroflow_2025}; and (3) open-source agentic models, including WebSailor-32B~\cite{li2025websailornavigatingsuperhumanreasoning}, WebDancer-QwQ~\cite{wu2025webdancerautonomousinformationseeking}, WebShaper-32B~\cite{tao2025webshaperagenticallydatasynthesizing}, DeepMiner-32B-RL~\cite{tang2025deepminer}, AFM-32B-RL~\cite{li2025chainofagentsendtoendagentfoundation}, Tongyi DeepResearch-30B~\cite{li2025tongyi}, and MiroThinker-v1.0-30B~\cite{bai2025mirothinker}.

\paragraph{Evaluation Benchmarks.}
We evaluate all models' performance on a broad set of agentic benchmarks covering both deep search and deep research scenarios. Specifically, the deep search benchmarks include BrowseComp~\cite{wei2025browsecomp}, GAIA~\cite{mialon2023gaia}, XBench-DeepSearch~\cite{xbench2025}, WebWalkerQA~\cite{wu2025webwalker}, FRAMES~\cite{krishna-etal-2025-fact}, and SEAL-0~\cite{pham2025sealqaraisingbarreasoning}. For deep research evaluation, we use Deep Research Bench RACE~\cite{du2025deepresearchbenchcomprehensivebenchmark}, which evaluates long-form, open-ended research reports and reports both an overall score and four fine-grained criteria: Comprehensiveness, Insight, Instruction Following, and Readability.

\paragraph{Metrics.}
We use LLM-as-judge approach for evaluation. For deep search tasks, we adopt the pass@1 metric with a specific judge prompt. For Deep Research Bench RACE, which assesses report quality across four criteria: Comprehensiveness (coverage of key aspects), Insight/Depth (analytical novelty), Instruction-Following (adherence to query constraints), and Readability (clarity and coherence), we use another judge prompt. See the two judge prompts in \ref{app:judge_deepresearch}.

\paragraph{Implementation Details.}
We conduct experiments using the backbone model: Qwen3-30B-A3B-Instruct-2507. During supervised fine-tuning, we train the models for 3.5 epochs with a batch size of 128, using the AdamW optimizer and a cosine decay learning rate schedule with an initial learning rate of $1.4\times10^{-5}$. The maximum sequence length is set to 65{,}536 tokens to support long-horizon trajectories. All models are trained under the same agentic workflow and data settings described in earlier sections. In the RL stage, the learning rate is set to $1\times10^{-6}$ with a batch size of 32. For each question, 8 on-policy rollouts are generated, with a maximum sequence length of 128k tokens, up to 120 interaction turns, and training is performed for 60 steps. During inference, we use vLLM, with a context window of 128K tokens. Unless otherwise specified, all experiments are conducted with a maximum of 100 interaction steps, a plan refinement interval of N=5 interaction steps.

\paragraph{Context Management during Inference Stage.}

Long-horizon tasks (e.g., BrowseComp) often exceed the effective context capacity of a vanilla agent under a 128K window, and this issue is amplified in SMTL because each interaction step produces more tool observations, reducing the number of steps that can be accommodated before hitting the context limit. To improve context efficiency, SMTL couples periodic plan refinement with an overflow-triggered compression scheme: the agent refines the task plan every $N{=}5$ steps by default, and when the accumulated history reaches the 128K context budget without a confirmed answer, it performs an additional forced plan refinement using the current history, then drops all pre-plan context and continues execution from the refreshed plan. This plan-centric reset preserves the latest execution state and subtask structure, keeping inference behavior aligned with training-time plan refinement. As a result, SMTL supports longer effective trajectories under a fixed context budget without sacrificing structured task context. To align with the interaction budgets commonly used by baselines, we further evaluate SMTL with maximum step limits of 100 and 300, referred to as SMTL-100 and SMTL-300, respectively. For Deep Research Bench, we report results under the 100-step setting, as open-ended research tasks typically converge within tens of interactions rather than exhausting the maximum interaction budget.

\begin{table*}[!ht]
\centering
\footnotesize
\definecolor{rowhl}{RGB}{220,245,225}
\caption{Main results on Deep Search and Deep Research benchmarks.
For Deep Search benchmarks, BC: BrowseComp; Xbench-DS: Xbench-DeepSearch; WW: WebWalker-QA; DR-Gym: DeepResearch Gym; DR-Bench: DeepResearch Bench. For Deep Research Bench, Comp., Depth, Inst., and Read. denote Comprehensiveness, Insight/Depth, Instruction Following, and Readability, respectively. We report pass@1 for our models. Results marked with $^{*}$ are obtained by our own evaluation, while results marked with $^{+}$ are taken from prior work~\citep{yao2026oresearcher}.}

\label{tab:main}
\vspace{-6pt}
\renewcommand\arraystretch{1.25}
\setlength{\tabcolsep}{3.5pt}
\resizebox{.95\textwidth}{!}{%
\begin{tabular}{l cccccc !{\hspace{2pt}\vrule\hspace{2pt}} ccccc} 
\toprule
\multirow{2}{*}{\textbf{Model}} & \multicolumn{6}{c}{\textbf{Deep Search}} & \multicolumn{5}{c}{\textbf{Deep Research Bench RACE}} \\
\cmidrule(lr){2-7}\cmidrule(lr){8-12}
 & \textbf{BC} & \textbf{GAIA} & \textbf{Xbench-DS} & \textbf{WW} & \textbf{FRAMES} & \textbf{SEAL-0} & \textbf{Overall.} & \textbf{Comp.} & \textbf{Depth} & \textbf{Inst.} & \textbf{Read.} \\
\midrule
\multicolumn{12}{c}{\cellcolor[RGB]{225, 235, 245}{\footnotesize\textit{Foundation Models with Tools}}} \\
GLM-4.5 & 26.4 & 66.0 & 70.0 & 65.6 & 78.9 & 36.0 & -- & -- & -- & -- & -- \\
Minimax-M2 & 44.0 & 75.7 & 72.0 & 64.5 & -- & -- & 46.1 & 45.2 & 44.6 & 49.0 & 44.7 \\
DeepSeek-V3.2 & 40.1 & 63.5 & 71.0 & -- & 80.2 & 38.5 & -- & -- & -- & -- & -- \\
Kimi-K2-0905 & 14.1 & 60.2 & 61.0 & 63.0 & 58.1 & 25.2 & 44.5 & 42.8 & 39.7 & 50.8 & 46.0 \\ 
Claude-4.5-Sonnet & 19.6 & 71.2 & 66.0 & -- & 85.0 & 53.4 & 39.9 & -- & -- & -- & -- \\
GPT-5 & 54.9 & 59.4 & -- & 73.0 & 90.0 & 51.4 & 46.8 & 45.4 & 44.5 & 50.3 & 47.5 \\
Gemini-2.5-Pro & - & 60.2 & 56.0 & -- & -- & 19.8 & 35.1 & 34.1 & 29.8 & 41.7 & 37.2 \\
\midrule
\multicolumn{12}{c}{\cellcolor[RGB]{245,235,220}{\footnotesize\textit{Deep Research System}}} \\
OpenAI DeepResearch & 51.5 & 67.4 & 26.6 & -- & -- & -- & 47.0$^{+}$ & 46.9$^{+}$ & 45.3$^{+}$ & 49.3$^{+}$ & 47.1$^{+}$ \\
Gemini DeepResearch & 59.2 & -- & 50.0 & -- & -- & -- & 48.9$^{+}$ & 48.5$^{+}$ & 48.5$^{+}$ & 49.2$^{+}$ & 49.4$^{+}$ \\
Perplexity Deep Research & 22.0 & -- & -- & 67.0 & 83.0 & 38.7 & 42.3$^{+}$ & 40.7$^{+}$ & 39.4$^{+}$ & 46.4$^{+}$ & 44.3$^{+}$ \\
Kimi-Researcher & 26.9 & -- & 69.0 & -- & 78.8 & 36.0 & 44.6 & 45.0 & 42.0 & 47.1 & 45.6 \\
OAgents (Claude-3-7) & 22.2 & 66.7 & 54.5 & 53.0 & -- & -- & 50.8$^{+}$ & 50.4$^{+}$ & 51.2$^{+}$ & 50.3$^{+}$ & 49.4$^{+}$ \\
MiroFlow (GPT-5) & 33.2 & 82.4 & 72.0 & 52.6 & -- & -- & 44.9 & 44.6 & 49.3 & 45.6 & 46.1 \\
\midrule
\multicolumn{12}{c}{\cellcolor[RGB]{250,245,200}{\footnotesize\textit{Open-source Agentic Model}}} \\
WebSailor-32B & 10.5 & 53.2 & 53.3 & 60.5 & 69.8 & 16.2 & 32.4 & 28.6 & 22.7 & 43.7 & 37.5 \\
WebDancer-QwQ & 3.8 & 51.5 & 40.0 & 47.9 & -- & 20.7 & 35.9$^{*}$ & 33.0$^{*}$ & 28.5$^{*}$ & 44.8$^{*}$ & 40.8$^{*}$ \\
WebShaper-32B & 9.0 & 52.4 & 54.6 & 51.4 & -- & -- & 34.9 & 31.6 & 26.2 & 44.8 & 40.4 \\
DeepMiner-32B-RL & 33.5 & 58.7 & 62.0 & -- & -- & -- & -- & -- & -- & -- & -- \\
AFM-32B-RL & 11.1 & 55.3 & 52.0 & 63.0 & -- & -- & 35.8$^{*}$ & 32.7$^{*}$ & 30.2$^{*}$ & 38.5$^{*}$ & 40.9$^{*}$ \\
\midrule
NestBrowse-30B-A3B & 31.6 & {\bfseries 75.7} & 75.0 & -- & -- & -- & -- & -- & -- & -- & -- \\
Tongyi-DeepResearch-30B & 43.4 & 70.9 & 75.0 & 72.2 & {\bfseries 90.6} & -- & 45.7$^{+}$ & {\bfseries 44.7$^{+}$} & 44.2$^{+}$ & 48.8$^{+}$ & 44.2$^{+}$ \\
MiroThinker-v1.0-30B & 41.2 & 73.5 & 70.6 & 61.0 & 85.4 & 46.8 & -- & -- & -- & -- & -- \\
\midrule
\rowcolor{rowhl} {\bfseries SMTL-100} & 43.6 & 74.8 & 80.0 & 74.9 & 84.3 & 50.5 & {\bfseries 45.9} & 42.1 & {\bfseries 45.6} & {\bfseries 49.6} & {\bfseries 45.5}\\
\rowcolor{rowhl} {\bfseries SMTL-300} & {\bfseries 48.6} & {\bfseries 75.7} & {\bfseries 82.0} & {\bfseries 76.5} & 85.1 & {\bfseries 51.4} & - & - & - & - &  \\
\bottomrule
\end{tabular}%
}
\end{table*}

\subsection{Main Results}

\paragraph{SMTL exhibits consistent Pareto dominance across heterogeneous benchmarks.}

As shown in Table~\ref{tab:main}, SMTL-30B demonstrates strong performance across Deep Search benchmarks under different interaction budgets. With a moderate budget (SMTL-100), the model already achieves state-of-the-art performance among 30B-scale open-source agentic models on BrowseComp (43.6\%), slightly surpassing Tongyi-DeepResearch-30B (43.4\%) and clearly outperforming MiroThinker-v1.0-30B (41.2\%). It also reaches 78.0\% on XBench-DeepSearch and 74.9\% on WebWalker-QA. When the budget is increased to 300 steps, performance further improves, most notably on the long-horizon benchmark BrowseComp, where accuracy rises from 43.6\% to \textbf{48.6\%} (+5.0), substantially widening the gap over both Tongyi and MiroThinker. In contrast, gains on shorter-horizon tasks such as GAIA (74.8\% → \textbf{75.7\%}) and WebWalker (74.9\% → \textbf{76.5\%}) are comparatively modest, indicating that additional interaction budget primarily benefits deeper multi-step evidence aggregation. Moreover, as shown in Figure~\ref{fig:efficiency}, across interaction budgets from 50 to 300 steps on BrowseComp, SMTL consistently lies on the Pareto frontier of accuracy versus trajectory length, while mainstream baselines fall below this curve, confirming its superior efficiency-aware scaling on long-horizon search.

Beyond deterministic deep search, SMTL further generalizes to open-ended deep research evaluation. On DeepResearch Bench RACE, SMTL-100 achieves an overall score of \textbf{45.9\%}, with strong and balanced performance across Comprehensiveness (42.1\%), Insight/Depth (45.6\%), Instruction Following (49.6\%), and Readability (45.5\%). This surpasses representative open-source agentic baselines such as WebSailor-32B (32.4\%), WebDancer-QwQ (35.9\%), WebShaper-32B (34.9\%), and AFM-32B-RL (35.8\%), and slightly outperforms Tongyi-DeepResearch-30B (45.7\%) and Kimi-Researcher (44.6\%), establishing strong competitiveness among 30B-scale systems. This demonstrates that the same parallel search framework transfers from accuracy-driven benchmarks to long-form research synthesis without task-specific modification. Furthermore, as illustrated in Figure~\ref{fig:generalization}, SMTL consistently achieves leading average performance across heterogeneous benchmarks, confirming that its gains are not confined to a single task but reflect robust generalization across diverse objectives and evaluation criteria.

\paragraph{Why SMTL is more efficient?}
We analyze the efficiency advantage of SMTL through a qualitative case study comparing SMTL-30B with a representative deep-search baseline, MiroThinker-v1.0-30B, on a BrowseComp task. As illustrated in Figure~\ref{fig:case_study_illustration}, SMTL localizes the key entity within 8 assistant turns, whereas MiroThinker-v1.0 requires 16 turns to reach the same evidence.

This difference arises from fundamentally different search organization strategies. SMTL decomposes the task into multiple hypothesized subtasks and explores them in parallel, allowing the agent to rapidly surface high-signal evidence and to periodically re-plan subtasks based on intermediate observations. As a result, SMTL quickly converges on the correct search direction and allocates subsequent interactions to evidence verification. In contrast, MiroThinker-v1.0 follows a strictly sequential interaction pattern, in which only a single tool call is permitted per round. Information gathering therefore proceeds incrementally, requiring repeated query reformulation and delaying the discovery of key evidence.

This case study demonstrates that SMTL’s efficiency gains do not stem from deeper per-step reasoning, but from parallel subtask exploration and staged re-planning. By reorganizing search execution rather than extending reasoning depth, SMTL substantially reduces the number of interaction rounds required to localize critical information and complete the task.

\section{Analysis}
\subsection{Efficiency Evaluation}
\label{sec_efficiency}

\begin{table}[!ht]
\centering
\small
\caption{\textbf{Interaction efficiency on BrowseComp.} 
Average number of assistant steps, average tool calls per step, and task accuracy.}
\label{tab:efficiency_browsecomp}
\vspace{-6pt}
\begin{tabular}{lccc}
\toprule
\textbf{Model} & \textbf{Steps} & \textbf{Avg. Tool Calls} & \textbf{Acc.} \\
\midrule
Tongyi-DeepResearch-30B & 75.2  & 1.0 & 43.4 \\
MiroThinker-v1.0-30B    & 206.0 & 1.0 & 41.2 \\
\midrule
SMTL-100 & \textbf{60.4}  & 3.5 & 44.6 \\
SMTL-300 & 150.7 & 3.7 & \textbf{48.6} \\
\bottomrule
\end{tabular}
\end{table}

We evaluate the efficiency of SMTL from the perspective of interaction complexity, measured by the number of assistant steps required to solve a task. As shown in Table~\ref{tab:efficiency_browsecomp}, SMTL achieves a more favorable balance between interaction cost and task performance compared to representative baselines on BrowseComp. SMTL-100 attains 44.6\% accuracy with an average of 60.4 assistant steps, slightly outperforming Tongyi-DeepResearch-30B (43.4\%) while requiring fewer steps (75.2). The contrast with MiroThinker-v1.0-30B is even more pronounced: MiroThinker requires 206.0 steps to reach 41.2\% accuracy, whereas SMTL-100 achieves substantially higher accuracy with less than one-third of the interaction cost.

This efficiency is closely related to SMTL’s parallel execution mechanism. Unlike sequential systems that invoke a single tool per round, SMTL performs an average of 3.5 tool calls per step, enabling concurrent evidence acquisition across subtasks. By aggregating more information within each interaction round, SMTL increases information density per step and reduces redundant query reformulation, resulting in shorter yet more effective trajectories.

Increasing the interaction budget to SMTL-300 further improves accuracy to 48.6\%, with 150.7 average steps. While the trajectory length grows as expected, the additional steps translate into measurable performance gains rather than inefficient looping. Figure~\ref{fig:efficiency} further illustrates that across different interaction budgets, SMTL consistently lies on the Pareto frontier in the accuracy–step plane, confirming that its performance improvements are achieved with well-controlled interaction complexity.

\subsection{Ablations on Maximum Interaction Steps.}

\begin{figure}[!ht]
  \centering
  \begin{subfigure}[t]{0.48\linewidth}
    \centering
    \includegraphics[width=\linewidth]{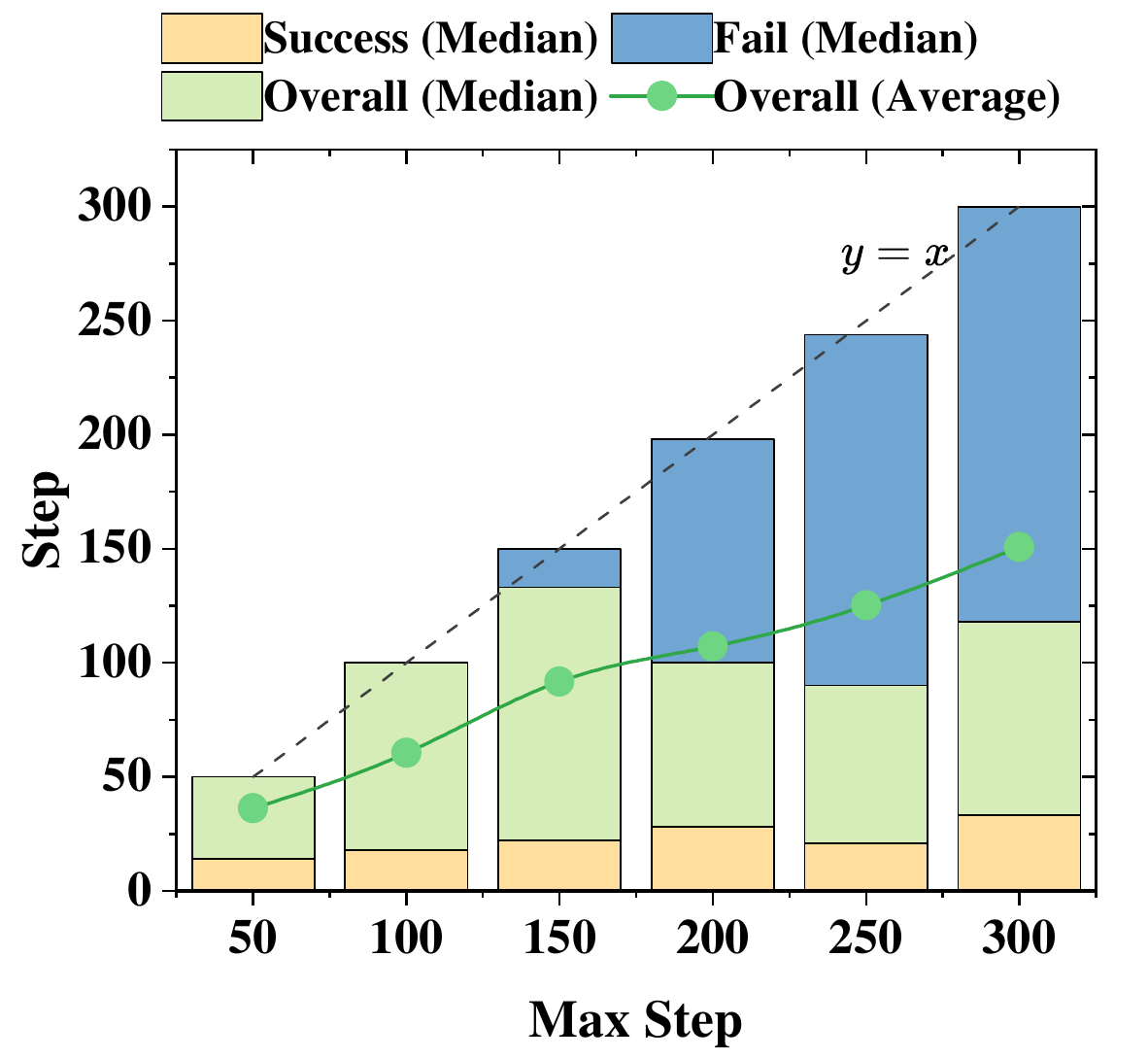}
    \vspace{-14pt}
    \caption{BC score under different max steps.}
    \label{fig:step_relationship}
  \end{subfigure}\hfill
  \begin{subfigure}[t]{0.48\linewidth}
    \centering
    \includegraphics[width=\linewidth]{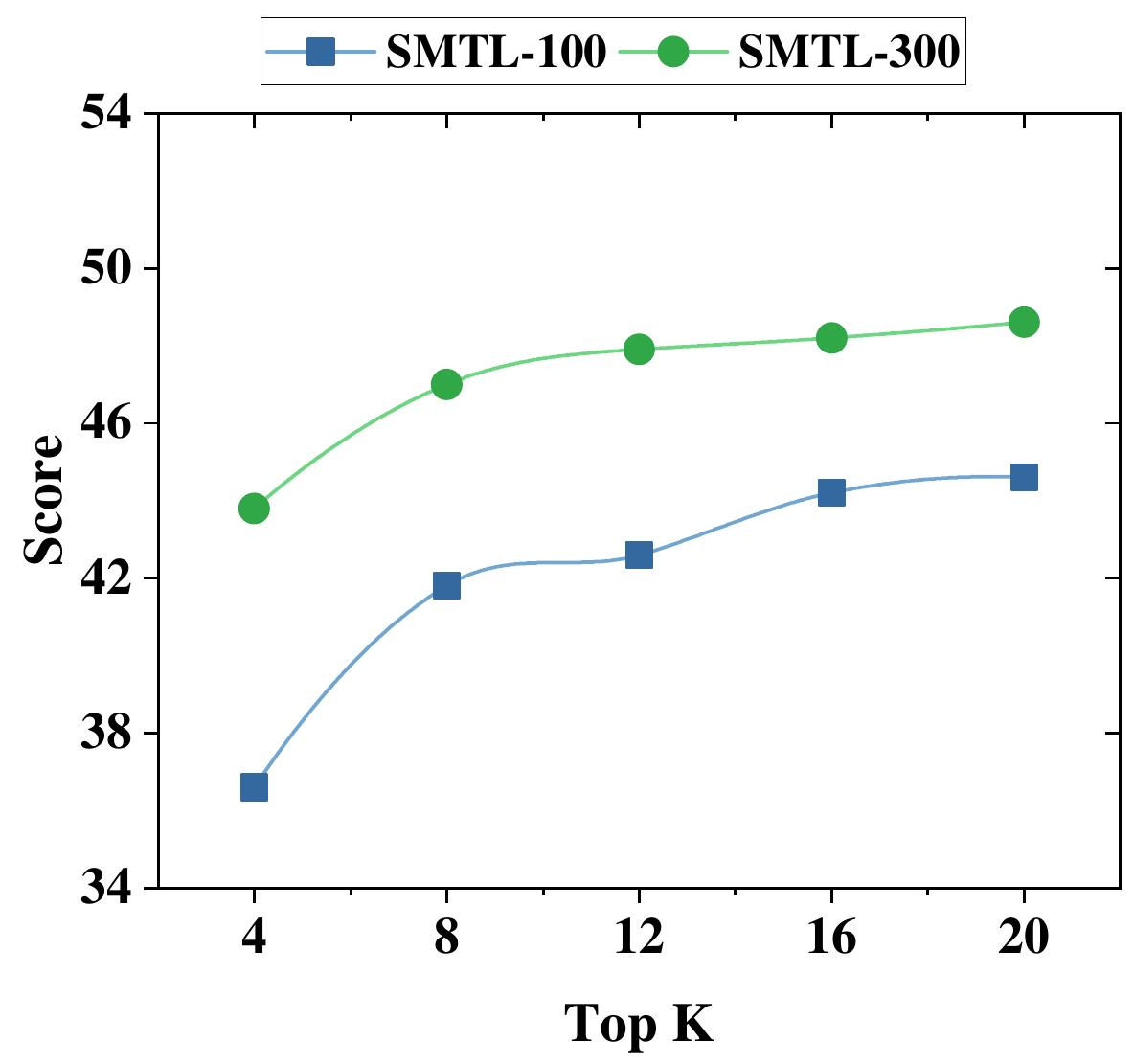}
    \vspace{-14pt}
    \caption{BC score under different web search topk results.}
    \label{fig:topk_relationship}
  \end{subfigure}
  \vspace{-4pt}
  \caption{Results of context management (CM) under different observation horizons $K$ with interaction budgets (IB) of 80 and 160 steps on the BC benchmark.}
  \label{fig:analysis}
\end{figure}

We conduct ablation studies to analyze how the maximum interaction budget (max steps) affects task outcomes and trajectory behavior in long-horizon agentic search. Specifically, we vary the maximum number of interaction steps from 50 to 300 on BrowseComp and report four statistics in Figure~\ref{fig:step_relationship}: overall average steps, overall median steps, the median steps of successful cases, and the median steps of failed cases.

Several clear patterns emerge. First, the median step count of successful cases does not exhibit a noticeable increasing trend as interaction steps grows. Most successful trajectories converge before reaching the interaction limit, suggesting that once a correct reasoning path is identified, additional budget provides limited benefit for these cases. In contrast, the median step count of failed cases closely follows the $y=x$ trend, indicating that the majority of failed trajectories terminate exactly at the maximum allowed step. This implies that many failures are due to exhausting the interaction budget rather than prematurely outputting incorrect answers. Consequently, the increase in overall average steps is primarily driven by the upward shift of failed cases, as more trajectories extend to the new budget ceiling before terminating. This observation suggests that the model is actively attempting to explore alternative reasoning paths when facing difficulty, rather than misunderstanding the task or exhibiting overconfidence through early answer generation.

We further analyze why increasing max interaction steps leads to performance gains. Under smaller budgets, a substantial portion of hard cases fail simply because SMTL cannot identify a valid reasoning path within the limited number of tool interactions. When the interaction budget is enlarged, SMTL is afforded additional opportunities to explore different evidence chains. Combined with periodic plan refinement, this extended budget enables our model to correct suboptimal search directions and progressively reorient toward promising subtasks. As a result, increasing max steps improves success rates primarily by alleviating search budget constraints in genuinely difficult cases, rather than by compensating for systematic reasoning errors.

\subsection{Ablations on Retrieval Top-$k$.}

We further investigate how the retrieval width of the web search tool affects performance by varying the parameter top-$k$, which controls the number of URLs returned for each query. We evaluate both SMTL-100 and SMTL-300 on BrowseComp under different top-$k$ settings.

As shown in Figure~\ref{fig:topk_relationship}, increasing top-$k$ consistently improves task performance. When top-$k$ increases from 4 to 8, both SMTL-100 and SMTL-300 exhibit substantial gains (e.g., SMTL-300 improves from 43.8 to 47.0, while SMTL-100 increases from 36.6 to above 41.8). This jump indicates that a narrow retrieval window significantly constrains evidence coverage, limiting the SMTL’s ability to identify relevant information within a fixed interaction budget. When top-$k$ further increases from 8 to 20, performance continues to improve, albeit at a slower rate and gradually converges. This suggests diminishing returns: once the most informative candidates are included, additional results contribute marginal gains but still enhance robustness by reducing the risk of missing critical evidence.

Overall, the results align with our design intuition that improving search breadth can be a powerful scaling dimension for long-horizon agentic search. Under a fixed number of interaction steps, increasing top-$k$ effectively packs more candidate evidence into each search action, raising the information density per step. Rather than extending reasoning depth, SMTL benefits more from broader evidence acquisition within each interaction, demonstrating that expanding retrieval breadth is a more efficient scaling axis for long-horizon search than merely increasing reasoning length.
\section{Conclusion}
In this work, we revisit long-horizon agentic search from the perspectives of efficiency and generalization. We propose \emph{Search More, Think Less} (SMTL), a unified agentic framework that replaces sequential reasoning with parallel evidence acquisition, enabling efficient long-horizon inference under constrained interaction budgets. SMTL integrates an efficient agentic workflow with structured context management and an automated data synthesis pipeline that supports both deterministic question answering and open-ended research tasks. Trained end-to-end with supervised fine-tuning and reinforcement learning, SMTL achieves state-of-the-art or competitive performance across a wide range of deep search and deep research benchmarks, while substantially reducing reasoning steps and inference latency. Our results demonstrate that prioritizing efficient, search-centric scaling over ever-deeper reasoning provides a practical and generalizable foundation for future deep research agents, and we hope this work encourages further exploration of efficiency-oriented agentic designs.

\clearpage

\newpage
\section{Contributions}

\textbf{Core Contributors}
\begin{tasks}[
            style=itemize,
            label-width=1em,
            column-sep=3.5em,
            before-skip=1.5ex,
            after-item-skip=1.5ex,
            ](2)
    \task Qianben Chen
    \task Tianrui Qin
    \task King Zhu
    \task Qiexiang Wang
    \task Chengjun Yu
    \task Shu Xu
    \task Jiaqi Wu
\end{tasks}

\textbf{Contributors}
\begin{tasks}[
            style=itemize,
            label-width=1em,
            column-sep=3.5em,
            before-skip=1.5ex,
            after-item-skip=1.5ex 
            ](2)
    \task Jiayu Zhang
    \task Xinpeng Liu
    \task Xin Gui
    \task Jingyi Cao
    \task Piaohong Wang
    \task Dingfeng Shi
    \task He Zhu
    \task Tiannan Wang
    \task Yuqing Wang
    \task Maojia Song
    \task Tianyu Zheng
    \task Ge Zhang
    \task Jian Yang
    \task Jiaheng Liu
    \task Minghao Liu
    \task Yuchen Eleanor Jiang
\end{tasks}

\textbf{Corresponding Author}
\begin{tasks}[
            style=itemize,
            label-width=1em,
            column-sep=3.5em,
            before-skip=1.5ex,
            after-item-skip=1.5ex 
            ](2)
    \task Wangchunshu Zhou
\end{tasks}

\clearpage
\bibliography{main}
\bibliographystyle{apalike}

\newpage
\appendix


\section{Tools Setup}
\label{app:tools}

Our agent operates with a minimal yet expressive toolset designed to support web-scale information acquisition and consolidation. Specifically, we employ two core tools: a search interface for candidate retrieval and a page-level crawler for content extraction and goal-directed summarization.

\begin{itemize}
    \item \textbf{\texttt{web\_search}.}  
    This tool provides access to web search functionality through the Serper API, which interfaces with the Google Search engine~\citep{Serper_2025_SerperAPI}. Given a model-generated query string, the tool retrieves a ranked list of search results, with the default setting returning the top five entries. Each result consists of a page title, a short snippet, and the corresponding URL. The search results serve as high-level signals for identifying potentially relevant sources and guiding subsequent crawling decisions.

    \item \textbf{\texttt{crawl\_page}.}  
    This tool is responsible for fine-grained content acquisition and structured summarization. It takes as input both a target URL and an explicit \emph{goal} describing the information need to be addressed. The URL is crawled using the Jina Reader API~\citep{JinaAI_2025_JinaReader}, after which the retrieved page content is summarized by the DeepSeek-V3.2 model~\citep{liu2025deepseekv32}.  
    Crucially, the goal specification provides semantic guidance for the summarization process, steering the model to extract and condense information that is directly relevant to the current subtask rather than producing a generic page summary. This goal-conditioned summarization enables more targeted evidence collection and reduces irrelevant context propagation. The prompt template used for page summarization is provided in Appendix~\ref{app:summary_prompt}.
\end{itemize}

\colorlet{punct}{red!60!black}
\definecolor{background}{HTML}{EEEEEE}
\definecolor{delim}{RGB}{20,105,176}
\colorlet{numb}{magenta!60!black}

\lstdefinelanguage{json}{
    basicstyle=\normalfont\ttfamily\small,
    numbers=left,
    numberstyle=\scriptsize,
    stepnumber=1,
    numbersep=8pt,
    showstringspaces=false,
    breaklines=true,
    frame=lines,
    backgroundcolor=\color{background},
    literate=
     *{0}{{{\color{numb}0}}}{1}
      {1}{{{\color{numb}1}}}{1}
      {2}{{{\color{numb}2}}}{1}
      {3}{{{\color{numb}3}}}{1}
      {4}{{{\color{numb}4}}}{1}
      {5}{{{\color{numb}5}}}{1}
      {6}{{{\color{numb}6}}}{1}
      {7}{{{\color{numb}7}}}{1}
      {8}{{{\color{numb}8}}}{1}
      {9}{{{\color{numb}9}}}{1}
      {:}{{{\color{punct}{:}}}}{1}
      {,}{{{\color{punct}{,}}}}{1}
      {\{}{{{\color{delim}{\{}}}}{1}
      {\}}{{{\color{delim}{\}}}}}{1}
      {[}{{{\color{delim}{[}}}}{1}
      {]}{{{\color{delim}{]}}}}{1},
}
\section{Data Construction}

\subsection{Deep Search Data}
\begin{tcolorbox}[
    colback=gray!5!white,
    colframe=gray!75!black,
    fonttitle=\bfseries,
    title=Entity-Centric Information Extraction Prompt,
    breakable
]

\textbf{Instruction:}  
Given structured entity metadata and a source content excerpt, extract entity-related information strictly based on explicit textual evidence.

\vspace{0.5em}
\textbf{Input Fields:}

\textbf{Entity Name:}
\begin{verbatim}
{entity_name}
\end{verbatim}

\textbf{Entity Type:}
\begin{verbatim}
{entity_type}
\end{verbatim}

\textbf{Entity Description:}
\begin{verbatim}
{description}
\end{verbatim}

\textbf{Source Content Excerpt:}
\begin{verbatim}
{source_content[:2000]}
\end{verbatim}

\vspace{0.5em}
\textbf{Task:}  
Extract entity-related information from the source content and return it in the following JSON format.

\begin{lstlisting}[language=json]
{
  "key_attributes": {
    // Extract attributes relevant to the entity type ({template})
    // Only include attributes that explicitly appear in the text
  },
  "surface_forms": [
    // Common surface mentions of the entity in the text
    // (e.g., abbreviations, short forms)
  ],
  "aliases": [
    // Fully equivalent alternative names
    // (e.g., full name, pen name, former name)
  ]
}
\end{lstlisting}

\vspace{0.5em}
\textbf{Extraction Constraints:}
\begin{enumerate}
    \item \textbf{Evidence-Only Extraction}:  
    Only extract information that explicitly appears in the provided source content.
    
    \item \textbf{No Inference or Fabrication}:  
    Do not infer, hallucinate, or supplement missing information.
    
    \item \textbf{Graceful Degradation}:  
    If a field cannot be extracted, return an empty object (\texttt{\{\}}) or empty list (\texttt{[]}).
\end{enumerate}

\vspace{0.5em}
\textbf{Output Requirements:}
\begin{itemize}
    \item The output must be a valid JSON object.
    \item Do not include any content outside the specified JSON fields.
\end{itemize}

\end{tcolorbox}

\begin{tcolorbox}[
    colback=gray!5!white,
    colframe=gray!75!black,
    fonttitle=\bfseries,
    title=Entity Description Factuality Evaluation Prompt,
    breakable
]

\textbf{Instruction:}  
Evaluate the factuality level of candidate entity descriptions and extract key factual information strictly based on the original text.

\vspace{0.5em}
\textbf{Factuality Evaluation Criteria and Scoring:}

\begin{itemize}
    \item \textbf{High Factuality (80--100 points)}:  
    The description contains concrete, objective facts such as numerical values, time, location, quantities, measurements, dates, or percentages.  
    \emph{Example keywords:} founded in 2010; 5,000 employees; headquartered in Beijing; market value of USD 10 billion; covers 500 square meters; November 2023.

    \item \textbf{Medium Factuality (50--79 points)}:  
    The description includes some specific information but is predominantly qualitative.  
    \emph{Example keywords:} large company; located in China; multiple services; rapid growth; industry participant.

    \item \textbf{Low Factuality (0--49 points)}:  
    The description is mainly abstract, subjective, or generic, lacking concrete data.  
    \emph{Example keywords:} important organization; high-quality services; influential; widely followed; has potential.
\end{itemize}

\vspace{0.5em}
\textbf{Candidate Entity List (Total: \{len(candidates)\}):}
\begin{verbatim}
{candidate_info}
\end{verbatim}

\vspace{0.5em}
\textbf{Evaluation Tasks (for \emph{each} entity):}
\begin{enumerate}
    \item Assign a factuality score between 0 and 100.
    \item Extract key factual information (\texttt{key\_info}) by directly copying the most factual phrases or words from the original description.  
    Multiple fragments should be separated by commas.
\end{enumerate}

\vspace{0.5em}
\textbf{Critical Constraint:}  
\emph{\texttt{key\_info} must be copied verbatim from the original description.}  
Do \textbf{not} paraphrase, rewrite, infer, or add new information.

\vspace{0.5em}
\textbf{Output Format:}  
Return the result as a valid JSON object with the following structure:

\begin{lstlisting}[language=json]
{
  "entity_scores": [
    {
      "entity_name": "Entity Name",
      "score": 85,
      "level": "High",
      "key_info": "founded in 2010, 5,000 employees, headquartered in Beijing"
    },
    {
      "entity_name": "Entity Name 2",
      "score": 45,
      "level": "Low",
      "key_info": "important organization, provides services"
    }
  ]
}
\end{lstlisting}

\vspace{0.5em}
\textbf{Additional Requirements:}
\begin{enumerate}
    \item \texttt{entity\_scores} must include all \{len(candidates)\} candidate entities.
    \item Preserve the original order of entities; no sorting is required.
    \item \texttt{level} must be one of: \texttt{"High"} (80--100), \texttt{"Medium"} (50--79), or \texttt{"Low"} (0--49).
    \item Entity names must exactly match those in the candidate list.
    \item \texttt{key\_info} must be copied directly from the original description.
    \item Scoring should be objective and unbiased; do not favor any entity.
\end{enumerate}

\end{tcolorbox}

\begin{tcolorbox}[
    colback=gray!5!white,
    colframe=gray!75!black,
    fonttitle=\bfseries,
    title=Hierarchical Fact-Centric Entity Description Generation Prompt,
    breakable
]

\textbf{Role Definition:}  
You are an expert responsible for generating a \emph{structured description} of a target entity based solely on objective, verifiable factual information.

\vspace{0.5em}
\textbf{Core Principle:}  
\emph{Use only objective, verifiable facts. Subjective, generic, or vague descriptions are strictly prohibited.}

\vspace{0.5em}
\textbf{Information to Be Extracted (Priority Order):}
\begin{enumerate}
    \item \textbf{Precise Numbers}: founding years, employee counts, revenue figures, rankings, IDs, award counts, etc.  
    \emph{Examples:} founded in 1959; more than 5{,}000 employees; annual revenue of 1 billion USD; won the Nobel Prize twice.
    
    \item \textbf{Specific Time}: exact dates, historical periods, event times.  
    \emph{Examples:} independence on December 12, 1963; participated in the Olympics in 2000; founded in the 1990s.
    
    \item \textbf{Specific Locations}: countries, cities, geographic positions, distances, elevations.  
    \emph{Examples:} located in East Africa; 5 km from the city center; 1{,}600 meters above sea level.
    
    \item \textbf{Specific Events}: historical events, awards, participations, published works.  
    \emph{Examples:} won the Nobel Prize in Literature; published ``XXX''; participated in the 2000 Olympics.
    
    \item \textbf{Measurable Attributes}: area, population, scale, institutional size.  
    \emph{Examples:} area of 580{,}000 square kilometers; population of 53 million; has 12 colleges.
\end{enumerate}

\vspace{0.5em}
\textbf{Strictly Prohibited Information:}
\begin{itemize}
    \item Subjective descriptions (e.g., important, famous, influential)
    \item Generic descriptions (e.g., provides services, is an organization)
    \item Vague expressions (e.g., many people, a certain region)
    \item Functional descriptions without concrete factual support
\end{itemize}

\vspace{0.5em}
\textbf{Key Requirements:}
\begin{itemize}
    \item All facts must be directly extracted from the provided descriptions and relations.
    \item Numerical, temporal, and locational information must appear \emph{verbatim} in the output.
    \item If a piece of information contains no objective facts, it must be ignored.
\end{itemize}

\vspace{0.5em}
\textbf{Special Emphasis: Hierarchical Abstraction Strategy}  
Present all extracted information hierarchically. Do not omit any factual information, and do not invent content.

\vspace{0.5em}
\textbf{Input Content:}
\begin{enumerate}
    \item \textbf{Current Entity Information:}
    \begin{itemize}
        \item Entity type: \texttt{\{entity\_type\}}
        \item Number of related items: \texttt{\{len(children\_descriptions)\}}
    \end{itemize}
    
    \item \textbf{Related Information List} (\texttt{\{len(children\_descriptions)\}} items):
\begin{verbatim}
{children_info_text}
\end{verbatim}

    \item \textbf{Relationships Among Related Information:}
\begin{verbatim}
{peers_info_text}
\end{verbatim}
\end{enumerate}

\vspace{0.5em}
\textbf{Output Content:}  
Generate the following three items:
\begin{enumerate}
    \item \textbf{facts}: a list of independent, objective, and verifiable facts extracted from the input.
    \item \textbf{summary}: a one-sentence abstraction of relationship types (e.g., work, geography, education).
    \item \textbf{question}: a multi-hop search question constructed strictly from the extracted objective facts.
\end{enumerate}

\vspace{0.5em}
\textbf{Entity Name Usage Rules:}
\begin{itemize}
    \item The name of the current entity must never be mentioned.
    \item Names of related entities must also never be mentioned in the question.
    \item Use abstract references such as ``an organization'', ``a country'', ``a company'', or relational descriptions.
\end{itemize}

\vspace{0.5em}
\textbf{Fact Extraction Procedure:}
\begin{enumerate}
    \item Scan related information for objective facts containing numbers, time, locations, or events.
    \item Categorize facts into numerical, temporal, locational, event-based, and measurement-based.
    \item Construct fact descriptions that integrate all objective details without using specific names.
    \item Ensure each fact contains at least one verifiable objective element.
\end{enumerate}

\vspace{0.5em}
\textbf{Question Construction Principles:}
\begin{itemize}
    \item The question must be built solely from extracted objective facts.
    \item All numerical, temporal, locational, event, and measurement information must be included.
    \item Completeness is more important than brevity.
    \item If insufficient objective facts exist, return an empty result.
\end{itemize}

\vspace{0.5em}
\textbf{Output Format:}
\begin{lstlisting}[language=json]
{
  "facts": [
    {
      "fact": "[single fact description without specific names]",
      "source": "child:0"
    },
    {
      "fact": "[single fact description without specific names]",
      "source": "child_peer:0_1"
    }
  ],
  "summary": "Abstract description of relationship types",
  "question": "Which {entity_type} satisfies the following conditions: [...]?"
}
\end{lstlisting}

\end{tcolorbox}

\begin{tcolorbox}[
    colback=gray!5!white,
    colframe=gray!75!black,
    fonttitle=\bfseries,
    title=Search-Enforced Entity Description Obfuscation Prompt,
    breakable
]

\textbf{Role Definition:}  
You are an expert responsible for obfuscating an entity description such that the correct answer \emph{cannot be inferred directly} and must be determined through external search.

\vspace{0.5em}
\textbf{Current Situation:}
\begin{itemize}
    \item \textbf{Entity type:} \texttt{\{entity\_type\}}
    \item \textbf{Entity name (ground truth):} \texttt{\{entity\_name\}}  
    \emph{(This name must never appear in the obfuscated description.)}
    \item \textbf{Original description:}
\begin{verbatim}
{description}
\end{verbatim}
\end{itemize}

\vspace{0.5em}
\textbf{Conditions in the Description:}
\begin{verbatim}
{facts_detail_text}
\end{verbatim}

\vspace{0.5em}
\textbf{Direct Reasoning Basis (to be neutralized):}
\begin{verbatim}
{reasoning}
\end{verbatim}

\vspace{0.5em}
\textbf{Entity Details:}
\begin{itemize}
    \item \textbf{Full description:}
\begin{verbatim}
{entity_description}
\end{verbatim}
    \item \textbf{Key attributes:}
\begin{verbatim}
{key_attrs_text}
\end{verbatim}
\end{itemize}

\vspace{0.5em}
\textbf{Strictly Prohibited Content:}
\begin{enumerate}
    \item Do not use prompt-instruction language (e.g., ``related information 0'').
    \item Do not mention the entity name \texttt{\{entity\_name\}} or any other specific entity name.
    \item Do not use expressions such as ``this \{entity\_type\}'' or ``the above \{entity\_type\}''.
\end{enumerate}

\vspace{0.5em}
\textbf{Required Natural Language Expressions:}
\begin{itemize}
    \item Use generic references (e.g., ``an organization'', ``a company'', ``a country'', ``a writer'').
    \item Use characteristic descriptions (e.g., ``a media company founded in year XX'').
    \item Use relational descriptions (e.g., ``the employing company'', ``the country of birth'').
\end{itemize}

\vspace{0.5em}
\textbf{Task:}  
Select \emph{one single most critical condition} to obfuscate such that the model can no longer directly infer the answer and must rely on search for verification.

\vspace{0.5em}
\textbf{Obfuscation Strategies:}
\begin{enumerate}
    \item \textbf{Time Obfuscation}:  
    Convert an exact time into a vague period or a search-dependent temporal description.  
    \emph{Examples:}  
    ``founded in 1949'' $\rightarrow$ ``founded in the late 1940s''.
    
    \item \textbf{Entity Obfuscation}:  
    Replace an entity name with search-required attributes.  
    \emph{Examples:}  
    ``founded by Steve Jobs'' $\rightarrow$ ``founded by a Silicon Valley entrepreneur whose name starts with `S'''.
    
    \item \textbf{Quantitative Obfuscation}:  
    Replace precise numbers with ranges or qualitative quantities.  
    \emph{Examples:}  
    ``won the Nobel Prize twice'' $\rightarrow$ ``won the Nobel Prize more than once''.
    
    \item \textbf{Information Deletion}:  
    Remove redundant or overly revealing conditions that make the answer trivial.
    
    \item \textbf{Concept Obfuscation}:  
    Abstract a specific concept into a higher-level description supported by searchable facts.
    
    \item \textbf{Combined Obfuscation}:  
    Apply multiple lightweight obfuscations while preserving coherence.
\end{enumerate}

\vspace{0.5em}
\textbf{Selection Criteria:}
\begin{itemize}
    \item Prioritize obfuscating the key condition used in direct reasoning.
    \item Choose the condition that most reduces certainty.
    \item Keep the description understandable and logically coherent.
    \item Ensure the answer remains discoverable via search.
    \item Avoid subjective or vague phrasing.
\end{itemize}

\vspace{0.5em}
\textbf{Output Format:}
\begin{lstlisting}[language=json]
{
  "target_fact_index": 0,
  "obfuscation_strategy": "Time obfuscation",
  "obfuscated_fact": "the obfuscated condition text",
  "obfuscated_description": "the full obfuscated description",
  "reasoning": "Explanation of why this condition was selected and how it was obfuscated"
}
\end{lstlisting}

\vspace{0.5em}
\textbf{Output Constraint:}  
Output \emph{only} the JSON object. Do not include any additional text.

\end{tcolorbox}

\begin{tcolorbox}[
    colback=gray!5!white,
    colframe=gray!75!black,
    fonttitle=\bfseries,
    title=Entity Answer Verification Prompt,
    breakable
]

\textbf{Instruction:}  
Answer the question strictly based on the provided description and identify the corresponding entity.

\vspace{0.5em}
\textbf{Input Fields:}

\textbf{Description:}
\begin{verbatim}
{description}
\end{verbatim}

\textbf{Question:}
\begin{verbatim}
What is this {entity_type}?
\end{verbatim}

\vspace{0.5em}
\textbf{Task:}
\begin{itemize}
    \item Infer the most specific entity name that satisfies the given description.
    \item Base your answer only on the provided description.
\end{itemize}

\vspace{0.5em}
\textbf{Output Format:}  
Return the result as a valid JSON object with the following fields:

\begin{lstlisting}[language=json]
{
  "answer": "Entity name",
  "reasoning": "Brief explanation of which conditions were used to infer the answer"
}
\end{lstlisting}

\vspace{0.5em}
\textbf{Constraints:}
\begin{itemize}
    \item You must return a \emph{specific and explicit} entity name.
    \item Do not respond with a concept, category, or abstract description.
    \item If the answer cannot be determined from the description, set \texttt{"answer"} to \texttt{"Cannot determine"}.
    \item Output \emph{only} the JSON object. Do not include any additional text.
\end{itemize}

\end{tcolorbox}

\begin{tcolorbox}[
    colback=gray!5!white,
    colframe=gray!75!black,
    fonttitle=\bfseries,
    title=Entity Answer Equivalence Verification Prompt,
    breakable
]

\textbf{Instruction:}  
Determine whether a predicted answer refers to the same entity as the ground-truth answer.

\vspace{0.5em}
\textbf{Input Fields:}

\textbf{Ground-Truth Answer:}
\begin{verbatim}
{entity_name}
\end{verbatim}

\textbf{Predicted Answer:}
\begin{verbatim}
{predicted_answer}
\end{verbatim}

\vspace{0.5em}
\textbf{Task:}
Evaluate whether the predicted answer correctly identifies the same entity as the ground-truth answer.

\vspace{0.5em}
\textbf{Output Format:}  
Return the result as a valid JSON object with the following fields:

\begin{lstlisting}[language=json]
{
  "is_correct": true,
  "explanation": "Brief explanation of why the answers match or do not match"
}
\end{lstlisting}

\vspace{0.5em}
\textbf{Judgment Criteria:}
\begin{itemize}
    \item If the two answers refer to the same entity (even if phrased differently), return \texttt{true}.
    \item If the predicted answer is \texttt{"Cannot determine"} or is clearly incorrect, return \texttt{false}.
    \item The predicted answer must be a \emph{specific and explicit entity name};  
    concepts, categories, or abstract descriptions are not acceptable.
    \item Consider valid aliases, abbreviations, and alternative names.
\end{itemize}

\vspace{0.5em}
\textbf{Output Constraint:}  
Output \emph{only} the JSON object. Do not include any additional text.

\end{tcolorbox}
\label{app:sec:search_data_construct}

\subsection{Deep Research Data}
\begin{tcolorbox}[
    colback=gray!5!white,
    colframe=gray!75!black,
    fonttitle=\bfseries,
    title=Graph-Grounded Open-Ended Research Question Construction Prompt,
    breakable
]

\textbf{Instruction:}  
Construct a high-quality open-ended research question strictly grounded in the provided subgraph structure and factual evidence.

\vspace{0.5em}
\textbf{Input Fields:}

\textbf{Target Entity:}
\begin{verbatim}
{target_entity}
\end{verbatim}

\textbf{Entity Type:}
\begin{verbatim}
{entity_type}
\end{verbatim}

\textbf{Supporting Subgraph:}
\begin{verbatim}
{subgraph_description}
\end{verbatim}

\vspace{0.5em}
\textbf{Task:}  
Generate \textbf{one open-ended research question} that requires synthesizing information from \emph{multiple nodes and relations} in the supporting subgraph.

The question should:
\begin{itemize}
    \item Require integrating evidence across the subgraph rather than relying on a single fact.
    \item Admit multiple valid answers supported by different reasoning paths.
    \item Be suitable for a \textbf{report-style response} involving explanation, comparison, or synthesis.
\end{itemize}

\vspace{0.5em}
\textbf{Construction Constraints:}
\begin{enumerate}
    \item \textbf{Graph-Grounded}:  
    The question must be answerable \emph{only} using information contained in the provided subgraph.
    
    \item \textbf{Non-Deterministic}:  
    Do not construct questions with a single exact or short factual answer.
    
    \item \textbf{Multi-Evidence Requirement}:  
    The question must require reasoning over multiple entities, attributes, or relations.
    
    \item \textbf{No Trivial Reformulation}:  
    Avoid paraphrasing factual descriptions or directly listing conditions.
\end{enumerate}

\vspace{0.5em}
\textbf{Output Format:}
\begin{verbatim}
{
  "research_question": "A single open-ended research question"
}
\end{verbatim}

\vspace{0.5em}
\textbf{Output Requirements:}
\begin{itemize}
    \item Output must be a valid JSON object.
    \item Do not include explanations, rationales, or additional text.
\end{itemize}

\end{tcolorbox}
\label{app:sec:research_data_construct}



\clearpage
\section{Case Study}
\begin{figure}[h]
\centering
\resizebox{\linewidth}{!}{%
\begin{tikzpicture}[
    font=\normalsize, 
    node distance=1.1cm and 1.2cm,
    box/.style={
        draw,
        rounded corners,
        align=left,
        inner sep=6pt,
        fill={gray!5!white},
        text width=#1
    },
    box/.default=6.2cm,
    lab/.style={font=\footnotesize, draw=none, fill=none, align=center},
    arrow/.style={->, thick}
]

\def\branchDrop{1.05cm}   
\def\branchShift{0.2cm}  

\node[box=13.2cm] (task) {
\textbf{Task:}\\
I'm looking for a historical figure. They had blue eyes and never drank alcohol. They married after moving to a different country from the one in which they had been born. They lost a child and wrote a letter asking for people to bring flowers to the child's resting place.\\
\textit{What was this person's name and title upon their accession to rulership?}
};

\node[box=3.6cm, below left=1.2cm and 0cm of task, align=center] (sm) {\textbf{SMTL-30B}};
\node[box=4.8cm, below right=1.2cm and 0cm of task, align=center] (mm) {\textbf{MiroThinker-v1.0-30B}};

\draw[arrow] (task.south) -- ++(0,-0.35) -| (sm.north);
\draw[arrow] (task.south) -- ++(0,-0.35) -| (mm.north);

\node[box=6.2cm, below=1.0cm of sm] (s1) {
\textbf{Stage 1: Parallel search over subtasks}\\
\textbf{Subtask 1:} Search clues: blue eyes; emigrated and married abroad; monarch\\
\textbf{Subtask 2:} Search clues: lost child; letter; flowers at resting place\\
\textbf{Subtask 3:} Search clues: blue eyes; emigrated and married abroad; never drank alcohol
};

\node[box=6.2cm, below=1.4cm of s1] (s2) {
\textbf{Stage 2: Re-planned subtasks}\\
\textbf{Subtask 1:} Search clues: blue eyes; abstained from alcohol\\
\textbf{Subtask 2:} Search clues: letter; flowers; child grave; royal context\\
\textbf{Subtask 3:} Search clues: royal figure; emigrated to another country; married abroad
};

\draw[arrow] (sm.south) -- (s1.north);
\draw[arrow] (s1.south) -- (s2.north);

\node[lab, right=0.15cm of s1] {Assistant\\turns 1--4};
\node[lab, right=0.15cm of s2] {Assistant\\turns 5--8};

\node[box=6.2cm, below=1.0cm of mm] (m1) {
\textbf{Stage 1:}\\
Search clues: flowers; child; grave; letter; ruler
};

\node[box=6.2cm, below=0.95cm of m1] (m2) {
\textbf{Stage 2}\\
Search clues: emigrated to another country; married abroad; ruler
};

\node[box=6.2cm, below=0.95cm of m2] (m3) {
\textbf{Stage 3}\\
Search clues: never drank alcohol
};

\node[box=6.2cm, below=0.95cm of m3] (m4) {
\textbf{Stage 4: Query reformulation}\\
Search clues: baby, grave, flowers, queen
};

\draw[arrow] (mm.south) -- (m1.north);
\draw[arrow] (m1.south) -- (m2.north);
\draw[arrow] (m2.south) -- (m3.north);
\draw[arrow] (m3.south) -- (m4.north);

\node[lab, left=0.15cm of m1] {Assistant\\turns 1--4};
\node[lab, left=0.15cm of m2] {Assistant\\turns 5--7};
\node[lab, left=0.15cm of m3] {Assistant\\turns 8--10};
\node[lab, left=0.15cm of m4] {Assistant\\turns 11--16};

\node[box=6.0cm, align=center, below=1.55cm of $(s2.south)!0.5!(m4.south)$] (key) {
\textbf{Key Entity Identified:}\\
\textbf{Queen Marie of Romania}
};

\draw[arrow] (s2.south) .. controls +(0,-0.8cm) and +(-1.0cm,0.8cm) .. (key.west);
\draw[arrow] (m4.south) .. controls +(0,-0.8cm) and +(1.0cm,0.8cm) .. (key.east);

\node[box=5.6cm, align=center,
      below left=\branchDrop and \branchShift of key] (sf) {
\textbf{Continue evidence verification}\\
Final answer produced at assistant turn 36
};

\node[box=5.6cm, align=center,
      below right=\branchDrop and \branchShift of key] (mf) {
\textbf{Continue evidence verification}\\
Final answer produced at assistant turn 150
};

\draw[arrow] (key.south west) -- (sf.north);
\draw[arrow] (key.south east) -- (mf.north);

\end{tikzpicture}%
}
\caption{Case study illustration comparing SMTL-30B and MiroThinker-v1.0-30B. SMTL performs parallel subtask execution with staged re-planning, enabling faster localization and verification of key evidence, while MiroThinker-v1.0-30B follows a strictly sequential search process.}
\label{fig:case_study_illustration}
\end{figure}
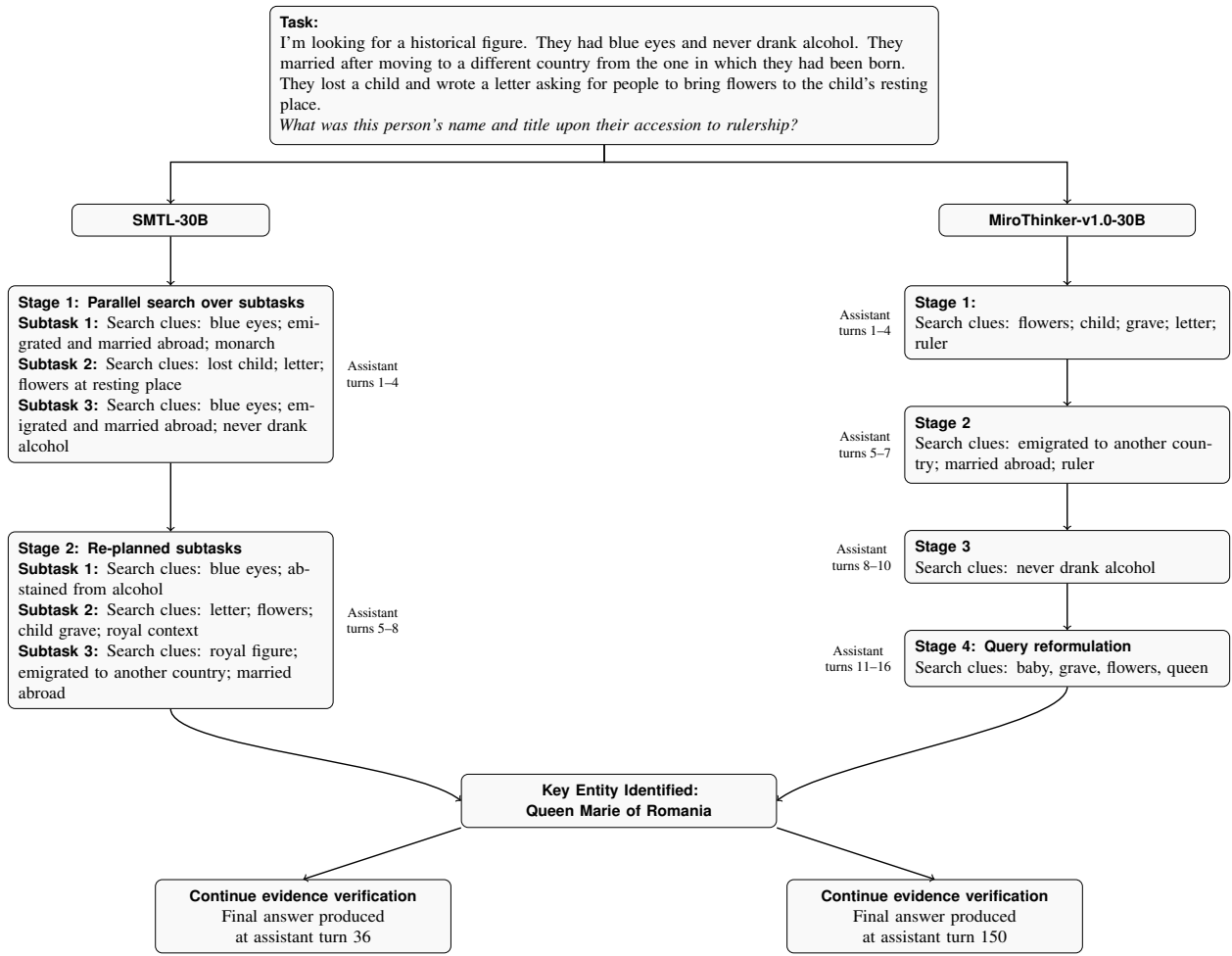

\section{Prompts}

\subsection{System Prompts}
\label{app:system_prompt}

We employ two system prompts to support \emph{Deep Search} and \emph{Deep Research} tasks, respectively. While the two prompts differ in their output structure and interaction protocols, they operate under a \emph{shared parallel agentic search framework} as described in the main paper.

Specifically, both system prompts follow a unified design philosophy: tasks are represented over graph-structured evidence, decomposed into multiple goals or subtasks, and solved through parallel execution and coordinated tool use. In both settings, the agent performs explicit planning, iterative plan refinement based on tool observations, and structured progress tracking, enabling efficient long-horizon search under constrained interaction budgets.

The primary distinction lies in the \emph{response format and organization}. The Deep Search prompt adopts a compact plan--plan-refine--answer structure optimized for deterministic question answering and efficient verification, whereas the Deep Research prompt enforces a finer-grained subtask-oriented protocol and report-style synthesis, tailored to open-ended research problems with multi-dimensional evaluation criteria. Despite these differences in output formatting and control flow granularity, both prompts instantiate the same underlying parallel agentic execution paradigm.

\begin{tcolorbox}[
    colback=gray!5!white,
    colframe=gray!75!black,
    fonttitle=\bfseries,
    title=System Prompt for Deep Search Tasks,
    breakable
]
\label{app:system_deepsearch}

\textbf{System Role:}  
You are an expert assistant that solves complex search tasks through structured tool usage and explicit goal management. You must follow a step-by-step execution process that combines planning, parallel execution, verification, and concise final answering.

\vspace{0.5em}
\textbf{Core Objective:}
\begin{itemize}
    \item Decompose the task into clear, verifiable goals.
    \item Advance multiple goals in parallel whenever possible.
    \item Execute each goal through sequential fallback paths.
    \item Use tools purposefully and efficiently to gather and verify evidence.
    \item Produce a final answer only after all goals are explicitly resolved.
\end{itemize}

\vspace{0.5em}
\textbf{Execution Requirements:}
\begin{enumerate}
    \item Follow a logical and structured order of tool usage.
    \item Parallelize independent goals; execute paths within a goal sequentially.
    \item Each action step must include:
    \begin{itemize}
        \item A brief reasoning explaining why the selected tool or path is chosen.
        \item One or more \texttt{<tool\_call>} executions with valid parameters.
        \item Interpretation of observations returned by tools.
        \item Refinement of subsequent actions based on observations.
    \end{itemize}
    \item Avoid redundant tool calls and repeated queries.
    \item Never assume task completion without explicit verification.
\end{enumerate}

\vspace{0.5em}
\textbf{Functional Interfaces:}

\textbf{1. \texttt{<plan>} Function}
\begin{itemize}
    \item \textbf{Role}: Decompose the task into parallelizable goals and execution paths.
    \item \textbf{Constraints}:
    \begin{itemize}
        \item 1--5 goals.
        \item Each goal has 1--5 sequential fallback paths.
        \item Each path must specify a clear success criterion.
    \end{itemize}
    \item \textbf{Timing}: Used only at the first step.
\end{itemize}

\textbf{2. \texttt{<plan\_refine>} Function}
\begin{itemize}
    \item \textbf{Role}: Summarize progress and decide next actions.
    \item \textbf{Content}:
    \begin{itemize}
        \item Plan recap.
        \item Execution status of each goal (Completed / In Progress / Blocked).
        \item Analysis of attempted paths.
        \item Next sub-paths to execute in parallel.
    \end{itemize}
    \item \textbf{Timing}: Invoked periodically after multiple actions.
\end{itemize}

\textbf{3. \texttt{<tool\_call>} Function}
\begin{itemize}
    \item \textbf{Available Tools}:
    \begin{itemize}
        \item \texttt{web\_search(query)}: Broad information discovery.
        \item \texttt{crawl\_page(url, query)}: Deep verification guided by an information goal.
    \end{itemize}
    \item \textbf{Usage Rules}:
    \begin{itemize}
        \item Use 1--5 tool calls per step, targeting distinct subtasks.
        \item Prioritize authoritative and official sources when precision is required.
        \item Always verify promising URLs via \texttt{crawl\_page}.
    \end{itemize}
\end{itemize}

\textbf{4. \texttt{<answer>} Function}
\begin{itemize}
    \item \textbf{Role}: Output the final confirmed answer.
    \item \textbf{Constraints}:
    \begin{itemize}
        \item Only after all goals are resolved.
        \item Must consolidate evidence across all goals.
        \item Language must match the task language.
        \item Keep the answer concise and factual.
    \end{itemize}
\end{itemize}

\vspace{0.5em}
\textbf{Critical Execution Rules:}
\begin{itemize}
    \item Advance all goals in parallel whenever possible.
    \item Never terminate early without verification.
    \item Do not skip promising sources during verification.
    \item Use no more than 10 tool calls per step.
\end{itemize}

\vspace{0.5em}
\textbf{Final Note:}  
Do not produce an answer unless you are absolutely certain. Final answers should be concise and avoid unnecessary explanation.

\end{tcolorbox}

\begin{tcolorbox}[
    colback=gray!5!white,
    colframe=gray!75!black,
    fonttitle=\bfseries,
    title=System Prompt for Deep Research Tasks,
    breakable
]
\label{app:system_deepresearch}
\textbf{System Role:}  
You are an expert assistant who solves tasks through structured tool calls, following a step-by-step process. Each step (action) involves analyzing needs, selecting tools, and executing calls to achieve the task. You are required to solve the task by formulating your thinking and reasoning process as described below.

\vspace{0.5em}
\textbf{Allowed Functions (Tag-based Protocol).}  
You can only use the following functions to answer a given question:
\texttt{subtask\_list}, \texttt{subtask}, \texttt{analysis}, \texttt{plan}, \texttt{tool\_call}, \texttt{tool\_response}, \texttt{subtask\_answer}, \texttt{answer}.

\begin{enumerate}
    \item \textbf{\texttt{subtask\_list}}: At the very beginning, break down the complex question into a list of clear, independent subtasks. Start with \texttt{<subtask\_list>} and end with \texttt{</subtask\_list>}.
    \item \textbf{\texttt{subtask}}: Marks the beginning of executing one specific subtask from \texttt{subtask\_list}. Start with \texttt{<subtask>} and end with \texttt{</subtask>}.
    \item \textbf{\texttt{analysis}}: Within a subtask, before using \texttt{plan} or \texttt{tool\_call}, provide reasoning, arguments, and next steps. Start with \texttt{<analysis>} and end with \texttt{</analysis>}.
    \item \textbf{\texttt{plan}}: For the current subtask, break it down into fine-grained steps to be executed using tools. Start with \texttt{<plan>} and end with \texttt{</plan>}.
    \item \textbf{\texttt{tool\_call}}: Execute tools from the tool list below to gather information relevant to answering the question.
    \item \textbf{\texttt{tool\_response}}: The response returned after a tool is executed.
    \item \textbf{\texttt{subtask\_answer}}: After completing a subtask, provide an intermediate, definitive answer. Start with \texttt{<subtask\_answer>} and end with \texttt{</subtask\_answer>}.
    \item \textbf{\texttt{answer}}: After all subtasks are completed, synthesize all \texttt{subtask\_answer}s into a final comprehensive answer.
\end{enumerate}

\vspace{0.5em}
\textbf{Available Tools.}
\begin{enumerate}
    \item \textbf{\texttt{web\_search}}: one parameter \texttt{query}. Example:
\begin{verbatim}
{'name': 'web_search', 'arguments': {'query': 'xxx'}}
\end{verbatim}
    \item \textbf{\texttt{crawl\_page}}: two parameters \texttt{url} and \texttt{query}. Example:
\begin{verbatim}
{'name': 'crawl_page', 'arguments': {'url': 'xxx', 'query': 'xxx'}}
\end{verbatim}
\end{enumerate}

\vspace{0.5em}
\textbf{Tool Usage Guide.}
\begin{enumerate}
    \item \textbf{\texttt{web\_search}}: If retrieved information is irrelevant, re-search with new queries until sufficient relevant information is obtained and you are highly confident in the final answer.
    \item \textbf{\texttt{crawl\_page}}: Use \texttt{crawl\_page} to obtain detailed information from URLs, and crawl additional URLs when needed.
    \item \textbf{Deeper search}: Use \texttt{web\_search} to discover URLs, then \texttt{crawl\_page} to verify and extract details. Repeat \texttt{web\_search}/\texttt{crawl\_page} multiple times if deeper hints emerge.
\end{enumerate}

\vspace{0.5em}
\textbf{Trail Notes.}
\begin{enumerate}
    \item \textbf{Overall workflow}: Start with \texttt{<subtask\_list>}. Then for each subtask, follow an \texttt{analysis $\rightarrow$ plan $\rightarrow$ tool $\rightarrow$ tool\_response} loop until the subtask is answerable, and output \texttt{<subtask\_answer>}. After all subtasks, produce \texttt{<answer>}.
    \item \textbf{Information gathering}: Based on the plan, you may use tools multiple times to collect sufficient external knowledge for the current subtask.
    \item \textbf{Tag restrictions}: \texttt{<subtask\_list>}, \texttt{<subtask>}, \texttt{<analysis>}, \texttt{<plan>}, \texttt{<web\_search>}, \texttt{<crawl\_page>}, \texttt{<tool\_response>}, \texttt{<subtask\_answer>}, \texttt{<answer>} are special tags and must not appear in free text, especially within the \texttt{<analysis>} function.
\end{enumerate}

\vspace{0.5em}
\textbf{Function Association Instructions.}
\begin{enumerate}
    \item The process must start with \texttt{<subtask\_list>}.
    \item After \texttt{<subtask\_list>}, the first \texttt{<subtask>} must begin.
    \item Inside each \texttt{<subtask>}, before using \texttt{plan} or tools, you must first use \texttt{<analysis>}.
    \item After information gathering is complete for a subtask, a \texttt{<subtask\_answer>} must be generated.
    \item Following a \texttt{<subtask\_answer>}, either start the next \texttt{<subtask>} or, if all are complete, generate \texttt{<answer>}.
\end{enumerate}

\vspace{0.5em}
\textbf{Answering Tips (Report Formatting).}  
The final answer must be a detailed, well-structured report with traceable information and must strictly follow:
\begin{enumerate}
    \item \textbf{Structure}: include an introduction, body paragraphs (organized by subtasks if appropriate), a conclusion, and a references section.
    \item \textbf{In-text citations}: every key information point, data point, or quote must be supported by a source. Use numbered square brackets with spaces, e.g., \texttt{ [1] }, \texttt{ [2] }, placed immediately after the supported claim.
\end{enumerate}

\vspace{0.5em}
\textbf{References Section.}  
Include a section titled \texttt{References} at the end of the report. It must be a numbered list corresponding one-to-one with citation numbers. Each entry must follow:
\begin{verbatim}
[Number]. URL - Webpage Title.
\end{verbatim}

\end{tcolorbox}

\subsection{Summary Prompt}
\label{app:summary_prompt}

\begin{tcolorbox}[
    colback=gray!5!white,
    colframe=gray!75!black,
    fonttitle=\bfseries,
    title=Goal-Conditioned Webpage Summarization Prompt,
    breakable
]

\textbf{Instruction:}  
Please process the provided webpage content and extract information that is directly relevant to the specified information goal.

\vspace{0.5em}
\textbf{Input Fields:}

\textbf{Webpage URL:}
\begin{verbatim}
{url}
\end{verbatim}

\textbf{Webpage Content:}
\begin{verbatim}
{content}
\end{verbatim}

\textbf{Information Goal:}
\begin{verbatim}
{query}
\end{verbatim}

\vspace{0.5em}
\textbf{Task Guidelines:}
\begin{enumerate}
    \item \textbf{Goal-Oriented Content Scanning}:  
    Identify the specific sections, passages, or data within the webpage that are directly relevant to the given information goal. Ignore unrelated or tangential content.

    \item \textbf{Evidence Extraction}:  
    Extract the most relevant information that supports the goal. Do not omit important details. Preserve the original wording and context as much as possible, and include extended excerpts when necessary (potentially spanning multiple paragraphs).

    \item \textbf{Structured Summarization}:  
    Organize the extracted information into a coherent and concise summary with a clear logical structure, prioritizing relevance, factual accuracy, and completeness with respect to the goal.
\end{enumerate}

\vspace{0.5em}
\textbf{Output Format:}  
The output must be a valid JSON object with the following fields:

\begin{lstlisting}[language=json]
{
  "rationale": "Explanation of why the extracted content is relevant to the information goal",
  "evidence": "Verbatim excerpts or minimally edited passages from the webpage that directly support the goal",
  "summary": "A concise, goal-focused summary synthesized from the extracted evidence"
}
\end{lstlisting}

\textbf{Additional Requirements:}
\begin{itemize}
    \item The output must be fully JSON-parsable.
    \item All special characters must be properly escaped.
    \item Do not include any content outside the specified JSON fields.
\end{itemize}

\end{tcolorbox}

\subsection{LLM-as-a-Judge Prompts}
\label{app:judge_prompts}

We design two distinct LLM-as-a-judge prompts for \emph{Deep Search} and \emph{Deep Research} evaluations, respectively. This separation is necessary because the two task settings produce fundamentally different types of outputs and therefore require different evaluation protocols to ensure fairness and reliability.

For Deep Search tasks, model outputs are typically short, deterministic answers with well-defined ground truth. Accordingly, the judge prompt focuses on semantic equivalence between the predicted answer and the labeled answer, allowing for minor surface-form variations while enforcing strict correctness. In contrast, Deep Research tasks produce long-form, report-style responses that involve synthesis, analysis, and organization across multiple sources. Evaluating such outputs requires a multi-dimensional rubric that assesses content coverage, analytical depth, instruction adherence, and readability rather than exact answer matching.

By tailoring the judge prompts to the structural and semantic characteristics of each task type, we ensure that evaluation remains consistent, unbiased, and aligned with the intended objectives of the benchmark. Both judge prompts are applied within the same evaluation framework but are specialized to match the output format and reasoning demands of their respective scenarios.

\begin{tcolorbox}[
    colback=gray!5!white, 
    colframe=gray!75!black, 
    fonttitle=\bfseries, 
    title=LLM-as-a-Judge Prompt for Deep Search Tasks, 
    breakable
]
\label{app:judge_deepsearch}
\textbf{Instruction:} Please determine whether the Predicted Answer is semantically equivalent to the Labeled Answer, given the Question.

\textbf{Input Format:}
\begin{verbatim}
Question: {question}
Labeled Answer: {gt_answer}
Predicted Answer: {pred_answer}
\end{verbatim}

\textbf{Evaluation Process:}
\begin{enumerate}
    \item \textbf{Instruction Focus}: Your judgment must be based solely on whether the meaning conveyed by the Predicted Answer aligns with the meaning of the Reference Answer. 
    \item \textbf{Allowable Variations}:
    \begin{itemize}
        \item For \textbf{text answers}: Differences in capitalization, punctuation, grammar (including prepositions, articles, and grammatical structures), word order, phrasing style, measurement units, or the inclusion/exclusion of non-essential descriptive phrases are \textbf{acceptable} if they do not alter the core meaning.
        \item For \textbf{names and titles}: Variations in name formats (e.g., inclusion of middle names, parentheses with additional information, honorifics) are acceptable if they refer to the same entity/person.
        \item For \textbf{numerical answers}: Minor acceptable margins of error are permissible when appropriate for the context.
        \item \textbf{Synonyms and near-synonyms} that convey the same meaning in context are acceptable.
    \end{itemize}
    \item \textbf{Criteria for CORRECT Judgement}: Judge as \textbf{"correct"} only if:
    \begin{enumerate}[label=\alph*.]
        \item The Predicted Answer \textbf{directly addresses} the Question, and
        \item Its \textbf{core meaning} is \textbf{semantically equivalent} to the Reference Answer, and
        \item It does \textbf{not contradict} any explicit requirements or constraints in the Question or Reference Answer.\\
        \textbf{Important}: If the Predicted Answer contains the Reference Answer within it (e.g., as an alternative name or with additional descriptive context) or expresses the same concept with different wording, it should be considered correct.
    \end{enumerate}
    \item \textbf{Criteria for INCORRECT Judgement}: Judge as \textbf{"incorrect"} if any of the following apply:
    \begin{enumerate}[label=\alph*.]
        \item The Predicted Answer \textbf{misses essential information} that changes the meaning of the Reference Answer (not just additional descriptive details), or
        \item The Predicted Answer \textbf{adds contradictory or incompatible information} that changes the intended meaning (mere additional non-contradictory details are acceptable), or
        \item The Predicted Answer is \textbf{ambiguous, indirect, or evasive} and fails to provide a clear, direct response to the Question, or
        \item The Predicted Answer is \textbf{semantically different} from the Reference Answer in a meaningful way that would lead to different understanding or action.
    \end{enumerate}
\end{enumerate}

\textbf{Special Considerations:}
\begin{itemize}
    \item \textbf{Preposition variations}: Differences in prepositions (e.g., "at" vs "in", "on" vs "upon") are generally acceptable unless they fundamentally change the meaning in the specific context.
    \item \textbf{Parenthetical information}: Inclusion of additional information in parentheses (e.g., alternative names, explanations) is acceptable if the core entity/answer remains the same.
    \item \textbf{Context sensitivity}: Consider the specific context of the question. For factual questions, focus on whether the same fact is conveyed.
\end{itemize}

\textbf{Output Format:}
\begin{lstlisting}[language=json]
{
  "rationale": "your rationale for the judgement",
  "judgement": "correct or incorrect"
}
\end{lstlisting}
\end{tcolorbox}

\begin{tcolorbox}[
    colback=gray!5!white,
    colframe=gray!75!black,
    fonttitle=\bfseries,
    title=LLM-as-a-Judge Prompt for Deep Research Tasks,
    breakable
]
\label{app:judge_deepresearch}

\textbf{System Role:}  
You are a strict, meticulous, and objective research article evaluation expert. You excel at using specific assessment criteria to deeply compare two articles on the same task, providing precise scores and clear justifications.

\vspace{0.5em}
\textbf{Task Background:}  
There is a deep research task, and you need to evaluate two research articles written for this task. The evaluation is conducted across four dimensions:
\textbf{Comprehensiveness}, \textbf{Insight}, \textbf{Instruction Following}, and \textbf{Readability}.

\textbf{Research Task:}
\begin{verbatim}
{task_prompt}
\end{verbatim}

\textbf{Articles to Evaluate:}

\textbf{Article 1:}
\begin{verbatim}
{article_1}
\end{verbatim}

\textbf{Article 2:}
\begin{verbatim}
{article_2}
\end{verbatim}

\textbf{Evaluation Criteria:}  
The articles should be evaluated and compared based on the following evaluation criteria list.  
Each criterion includes a detailed explanation and should be carefully considered.

\begin{verbatim}
{criteria_list}
\end{verbatim}

\textbf{Instruction:}

\textbf{Your Task}  
Please strictly evaluate and compare \texttt{<article\_1>} and \texttt{<article\_2>} based on \textbf{each criterion} in the \texttt{<criteria\_list>}. You must:

\begin{enumerate}
    \item \textbf{Analyze Each Criterion}: Assess how each article fulfills the requirements of the given criterion.
    \item \textbf{Comparative Evaluation}: Compare the performance of the two articles for each criterion, explicitly referencing the article content and the criterion explanation.
    \item \textbf{Score Separately}: Assign a score to each article for each criterion on a scale from 0 to 10.
\end{enumerate}

\textbf{Scoring Rules:}

\begin{itemize}
    \item \textbf{0--2}: Very poor performance; almost completely fails to meet the criterion.
    \item \textbf{2--4}: Poor performance; minimally meets the criterion with major deficiencies.
    \item \textbf{4--6}: Average performance; basically meets the criterion.
    \item \textbf{6--8}: Good performance; largely meets the criterion with notable strengths.
    \item \textbf{8--10}: Excellent performance; fully meets or exceeds the criterion.
\end{itemize}

\textbf{Output Format Requirements:}  
The output must \textbf{strictly follow} the JSON format below.  
Do \textbf{not} include any introduction, summary, or unrelated content.  
Evaluation should start from ``Standard 1'' and proceed sequentially through all criteria.

\begin{lstlisting}[language=json]
{
    "comprehensiveness": [
        {
            "criterion": "Text content of the first comprehensiveness evaluation criterion",
            "analysis": "Comparative analysis",
            "article_1_score": 0-10,
            "article_2_score": 0-10
        },
        ...
    ],
    "insight": [
        {
            "criterion": "Text content of the first insight evaluation criterion",
            "analysis": "Comparative analysis",
            "article_1_score": 0-10,
            "article_2_score": 0-10
        },
        ...
    ],
    ...
}
\end{lstlisting}
\textbf{Additional Requirement:}  
Ensure that the output JSON is fully parsable and that all characters that may cause JSON parsing errors are properly escaped.

\end{tcolorbox}

\end{document}